\documentclass[journal,10pt]{IEEEtran}
\usepackage[utf8, latin1]{inputenc} 
\usepackage{float} 
\usepackage{cite}
\ifCLASSINFOpdf
  \usepackage[pdftex]{graphicx}
  \DeclareGraphicsExtensions{.pdf,.jpeg,.png}
\else
  \usepackage[dvips]{graphicx}
  \DeclareGraphicsExtensions{.eps}
\fi

\usepackage{cite}
\usepackage{amssymb,amsfonts}
\usepackage[ruled]{algorithm2e}
\SetKwInput{KwReturn}{Return}
\SetKwInput{KwDescription}{Description}

\usepackage[cmex10]{amsmath}
\usepackage{algorithmic}
\usepackage{array}
\usepackage{algorithm2e}

\usepackage{fixltx2e}
\usepackage{stfloats}
\usepackage{url}
\usepackage{amssymb}
\usepackage{graphicx} 
\usepackage{gensymb} 
\usepackage{multirow}

\usepackage{booktabs} 
\usepackage{gensymb}
\usepackage{xspace}
\usepackage{nicefrac}

\usepackage[flushleft]{threeparttable}

\usepackage{caption}
\usepackage{subcaption}
\usepackage{color}
\DeclareMathOperator*{\argmax}{argmax}

\newcommand{\vect}[1]{\boldsymbol{\mathrm{#1}}}

\newcommand{\rvv}[1]{\textcolor{black}{#1}}
\newcommand{\rvvv}[1]{\textcolor{black}{#1}}

\makeatletter
\DeclareRobustCommand\onedot{\futurelet\@let@token\@onedot}
\def\@onedot{\ifx\@let@token.\else.\null\fi\xspace}

\def\ie{\emph{i.e}\onedot} 
 
\def\etc{\emph{etc}\onedot} \def\vs{\emph{vs}\onedot}
\def\wrt{w.r.t\onedot} 

\makeatother

\begin{document}
\title{\textit{TrAISformer}---A Transformer Network with Sparse Augmented Data Representation and Cross Entropy Loss for AIS-based Vessel Trajectory Prediction}

\author{\IEEEauthorblockN{Duong Nguyen, ~\IEEEmembership{Member,~IEEE},
and
Ronan Fablet, ~\IEEEmembership{Senior Member,~IEEE}}


\thanks{Duong Nguyen and Ronan Fablet are with IMT Atlantique, Lab-STICC, 29238 Brest, France (email: nvduong0512@gmail.com and ronan.fablet@imt-atlantique.fr}

\thanks{This work was supported by public funds (Minist\`ere de l'Education Nationale, de l'Enseignement Sup\'erieur et de la Recherche, FEDER, R\'egion Bretagne, Conseil G\'en\'eral du Finist\`ere, Brest M\'etropole). It benefited from HPC and GPU resources from Azure (Microsoft EU Ocean awards) and from GENCI-IDRIS (Grant 2020-101030).
The authors also acknowledge the support of ANR (French Agence Nationale de la Recherche) under reference ANR AI Chair OceaniX (ANR-19-CHIA-0016).}
}


\maketitle

\begin{abstract}

\rvv{Vessel trajectory prediction plays a pivotal role in numerous} maritime applications and services. While the Automatic Identification System (AIS) \rvv{offers} a rich source of information to address this task, forecasting vessel trajectory using AIS data remains challenging, even for modern machine learning techniques, because of the inherent heterogeneous and multimodal nature of motion data. In this paper, \rvv{we propose a novel approach to tackle these challenges. We introduce a discrete, high-dimensional representation of AIS data and a new loss function designed to explicitly address heterogeneity and multimodality}. 
The proposed model---referred to as \textit{TrAISformer}---is a modified transformer network that extracts long-term \rvv{temporal patterns in} \rvv{AIS vessel trajectories} in the proposed enriched space to forecast the positions of vessels several hours ahead. We report experimental results on real, publicly available AIS data. \textit{TrAISformer} significantly outperforms state-of-the-art methods, with an average prediction performance below 10 nautical miles up to $\sim$10 hours.

\end{abstract}


\section{Introduction}
\label{sec:introduction}

In
the last decades, the development of maritime activities has \rvv{raised} significant concerns relating to Maritime Surveillance (MS) and Maritime Situational Awareness (MSA), with vessel trajectory prediction being a focal point. Anticipating the direction of vessels and their approximate locations \rvv{at medium-range time horizons, ranging from a few tens of minutes to tens of hours ahead}, is at the core of diverse MS and MSA applications, including but not limited to search and rescue \cite{ou_ais_2008,varlamis_detecting_2018}, traffic control \cite{fabbri_optimization_2015}, \rvvv{path planning \cite{tu_exploiting_2017, dang_path_2022, nguyen_robust_2023}}, port congestion avoidance \cite{mou_study_2010, liu_ais_2020, kang_study_2021}, pollution monitoring \cite{soldi_space-based_2021}.

\begin{figure}
  \centering
  \includegraphics[width=\linewidth]{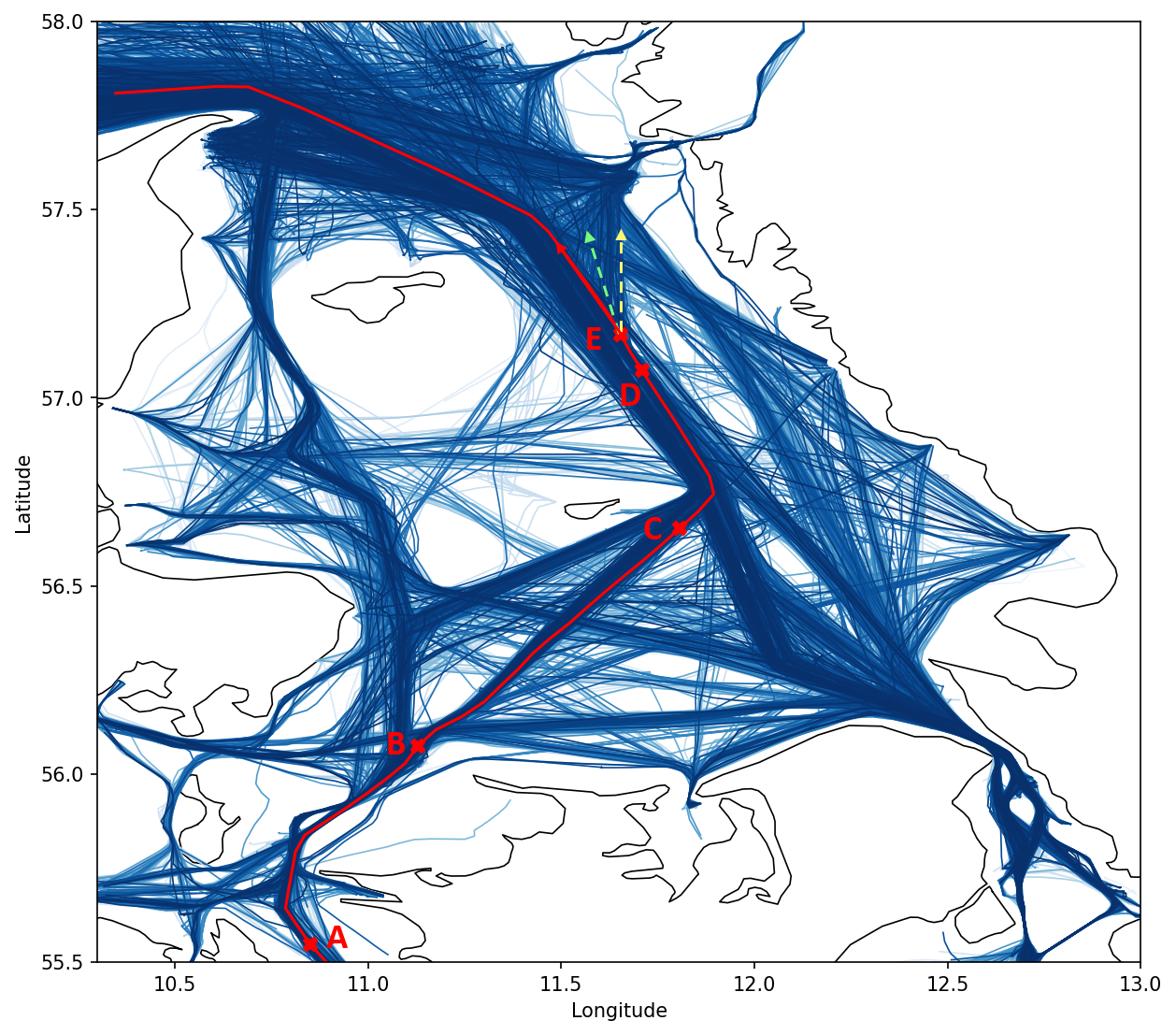}
  \centering
  \caption{{\bf Illustration of long-term\protect\footnotemark dependence patterns in \rvv{AIS vessel trajectories}:} At \textbf{E}, vessels typically follow one of the two main maritime routes indicated by the red and the yellow dashed arrows.  In order to forecast whether a vessel will continue straight ahead (the red path) or turn right (the yellow path), the prediction model may need to roll back several time steps to \textbf{D}, \textbf{C}, \textbf{B}, and \textbf{A} to understand the vessel's previous movements. Moreover, if the prediction model is not multimodal, it may output as a prediction an unusual green dashed path, \rvv{which is a merged path of} the true red and yellow ones.} \label{fig:maritmeTrafficGraph}
\end{figure}

The Automatic Identification System---AIS provides invaluable information for the monitoring and surveillance of maritime traffic. AIS data provide vessels' kinetic information (the current position indicated by the latitude and longitude coordinates, the current Speed Over Ground---SOG, the current Course Over Ground--COG, \etc), the information of the voyages, as well as the static information (the identification number in the format a Maritime Mobile Service Identity---MMSI number, the name of the vessel, \emph{etc.}) of vessels in the vicinity. Vessel trajectory prediction using AIS data has been studied for more than a decade \cite{ristic_statistical_2008, mazzarella_knowledge-based_2015, rong_ship_2019, forti_prediction_2020, volkova_predicting_2021, murray_ship_2021, park_ship_2021, capobianco_deep_2021}. However, the achievements thus far have been still limited. Most state-of-the-art schemes reach a relevant prediction performance only for short time horizons (ranging from a few minutes to half an hour) \cite{forti_prediction_2020, murray_ais-based_2021}, or for longer time horizons under particular movement patterns corresponding to predefined maritime routes \cite{millefiori_validation_2015, millefiori_modeling_2016} or unimodal movement patterns \cite{capobianco_deep_2021, murray_ais-based_2021}. Due to the complexity of vessel movement patterns and the heterogeneous nature of AIS data, forecasting vessel positions above several hours remains highly challenging.
As an illustration, we report in Fig. \ref{fig:maritmeTrafficGraph} two vessel paths with very similar movement patterns on segment \textbf{C}$\rightarrow$\textbf{D} but heading to two different destinations.

\footnotetext{\rvv{In this paper, we use the terms ``medium-range time horizon'', ``medium-range forecasting'', \etc to indicate the prediction horizons ranging from a few tens of minutes to tens of hours. The terms ``long-term dependency'', ``long-term correlation'', \etc., on the other hand, indicate the correlations across several time steps in the series.}}

Trajectory prediction has gained significant attention in recent years, particularly in the context of pedestrian and vehicle movement patterns \cite{rudenko_human_2020, zhao_tnt_2021, leon_review_2021, gu_densetnt_2021}. In this context, deep learning schemes have emerged as the state-of-the-art approach \cite{gupta_social_2018, vemula_social_2018, sadeghian_sophie_2019, zhao_multi-agent_2019, salzmann_trajectron_2021}. These recent advances however barely transfer to \rvv{AIS-based vessel trajectory} prediction \cite{liu_stmgcn_2022, liu_deep_2022}. First, the targeted space-time scales strongly differ (e.g.,  meters and a few seconds to a few minutes for pedestrian movements vs. kilometers and hours for vessel movements). Second, while interaction factors are critical to understanding and predicting pedestrian and vehicle trajectories, they have negligible effects on vessels' movements in the open sea. Besides, long-term dependencies are key factors for the latter and need to be explicitly addressed in \rvv{AIS-based vessel trajectory} prediction models.

In the open sea, vessels often follow some common movement patterns in order to optimize fuel consumption and to conform with maritime traffic regulations\cite{coscia_multiple_2018, varlamis_network_2019}. However, the analysis of the maritime traffic according to a finite set of interconnected maritime routes using clustering-based approaches \cite{millefiori_validation_2015, millefiori_modeling_2016, capobianco_deep_2021, murray_ais-based_2021} appears too simplistic to account for the heterogeneous and multimodal characteristics of real-world AIS data. \rvv{The core challenge of vessel trajectory prediction for the targeted time horizon in this paper (ranging from half an hour to tens of hours) revolves around accurately predicting the turning direction at the ``intersections"---commonly referred to as \textit{waypoints}---along maritime routes (see Fig. \ref{fig:maritmeTrafficGraph}). From a mathematical perspective, this involves developing a method that effectively represents the multimodal nature of AIS data at these waypoints, where each turning direction corresponds to a distinct mode of the data distribution. In order to forecast the trajectory correctly, the prediction model may need to backtrack several time steps to know where the vessel comes from and to fully understand large-scale movement patterns.  In essence, we can identify two primary methodological challenges: i) learning how to represent maritime traffic flows beyond a fully-structured network of maritime routes; and ii) capturing multi-scale patterns in vessel trajectories.} 

To address these challenges, we propose a novel deep learning model, referred to as \textit{TrAISformer}.  Our key contributions are as follows:
\begin{itemize}
    \item \rvv{\textit{TrAISformer} exploits a specific sparse, high-dimensional representation of AIS data and frames the prediction task as a classification problem to explicitly model the heterogeneity of AIS data and the multimodality of vessel trajectories.
    \item We leverage a probabilistic transformer architecture to capture long-term dependencies in \rvv{AIS vessel trajectories}.}
    \item We benchmark \textit{TrAISformer} w.r.t. state-of-the-art schemes on a real AIS dataset and report a prediction error below 10 nmi (nautical mile) up to 10 hours, which significantly outperforms previous works \cite{forti_prediction_2020, murray_ais-based_2021, capobianco_deep_2021}.
\end{itemize}

The paper is organized as follows. Section \ref{sec:related_work} states the problem and gives an overview of the related work and current limitations for \rvv{AIS-based vessel trajectory} prediction. We present the proposed approach in Section \ref{sec:proposedApproach}. Section \ref{sec:experiments} details our numerical experiments. We further discuss our main contributions and future work in Section \ref{sec:conclusions}.
 
\section{Problem statement and related work}
\label{sec:related_work}

\rvv{AIS-based vessel trajectory} prediction involves forecasting the future positions of vessels over a specific time horizon, using a series of historical AIS data. Formally, let us denote by $\vect{x}_t$ an AIS observation at time step $t$, where $\vect{x}_t$ comprises the position of the vessel (indicated by the latitude and the longitude coordinates), its Speed Over Ground---SOG, and its Course Over Ground---COG\footnote{We let the reader refer to \cite{bosnjak_automatic_2012} for a more detailed presentation of AIS data streams.}. 
\begin{equation}
    \vect{x}_t \triangleq \left[lat, lon, SOG, COG \right]^T.
    \label{eq:4d_representation}
\end{equation}
An {AIS vessel trajectory} is then represented by a series of observations $\{ \vect{x}_{t_0}, \vect{x}_{t_1}, ...\vect{x}_{t_T} \}$ where $t_i < t_j$ if $i < j$. One can use some simple interpolation method to get a series of $T+1$ equally sampled observations $\vect{x}_{0:T} \triangleq \{ \vect{x}_{t_0}, \vect{x}_{t_0+\Delta t}, \vect{x}_{t_0+2*\Delta t},..., \vect{x}_{t_0+T*\Delta t} \}$. The time step $\Delta t$ is chosen such that the error inherited from the interpolation has a negligible effect on downstream tasks. In this paper, we fix the time step at 10 minutes and omit $\Delta t $ for the sake of notation simplicity, \ie $\vect{x}_{t_0+n*\Delta t} \triangleq \vect{x}_{t+n}$.  

The $L$-step-ahead prediction problem comes to predict the trajectory $\vect{x}_{T+1:T+L} \triangleq  \{\vect{x}_{T+1}, \vect{x}_{T+2},...,\vect{x}_{T+L}\}$ given $T+1$ observations $\vect{x}_{0:T} \triangleq \{\vect{x}_0, \vect{x}_1,..., \vect{x}_T\}$ up to time $T$. Given the nature of the prediction problem, it naturally arises as the sampling of the following conditional distribution:
\begin{equation}
    p(\vect{x}_{T+1:T+L}|\vect{x}_{0:T}).
    \label{eq:obj_basic}
\end{equation}
We may point out that this probabilistic formulation also covers deterministic prediction models \cite{forti_prediction_2020, suo_ship_2020}, which reduce $p$ to Dirac distributions.

\rvv{There are two primary categories of approaches to \rvv{AIS-based vessel trajectory} prediction. The first one relies on a state-space formulation, which combines a dynamical prior on vessel movements with a filtering method to infer or sample the posterior \eqref{eq:obj_basic}. Models following this approach have certain limitations. Firstly, the dynamical prior often relies on a simplistic model, such as the Curvilinear Motion Model (CMM) \cite{perera_maritime_2012}, which cannot account for complex patterns including turning points. Secondly, filtering methods, such as the Kalman filter \cite{perera_maritime_2012, fossen_extended_2018} or the particle filter \cite{ristic_statistical_2008}, are prone to error propagation issues. Consequently, they seem less suitable for medium-range prediction.}

\rvv{In recent years, the second approach, known as the learning-based approach, has gained substantial popularity \cite{zhang_vessel_2022}. \cite{wang_vessel_2020}, \cite{li_long-term_2019}, \cite{tang_model_2019} leveraged LSTM (Long Short-Term Memory) and GRU (Gated Recurrent Unit) to learn the temporal patterns in $\vect{x}_{0:T}$. However, given the multi-path patterns exhibited in AIS data, such schemes are likely to fail \cite{nguyen_multi-task_2018}. More sophisticated models take into account the interactions between vessels. \cite{liu_deep_2022} used a customized pooling layer---referred to as Collision-Free Social Pooling (CFSP), while \cite{liu_stmgcn_2022} 
employed Graph Convolutional Neural Networks (GCN) to model the interactions between vessels in proximity. These models demonstrate improved prediction performance in dense traffic scenarios with relatively short prediction horizons, typically below one hour. However, in the open sea, where vessel density is significantly lower, and for medium-range horizons (spanning from a few hours to tens of hours), the impacts of the interactions between vessels are less pronounced. Furthermore,  since the surrounding environment of a vessel may change as vessels enter or exit the considered zone, it is intractable to explicitly model such interactions for these time horizons (see the Appendix).}

\rvv{To address the heterogeneous and multimodal nature of AIS data, several methods rely on clustering \cite{pallotta_vessel_2013, mazzarella_knowledge-based_2015, forti_prediction_2020, capobianco_deep_2021, murray_ais-based_2021}. They assume that maritime traffic in a given area can be represented as a graph, where each node corresponds to a \textit{waypoint} and each edge represents the maritime route between two nodes. The prediction problem then resorts to exploiting a forecasting model over the defined graph. A rich literature exists and exploits among others the constant velocity model and the particle filter \cite{mazzarella_knowledge-based_2015}, Gaussian Processes \cite{rong_ship_2019} and neural networks \cite{murray_ais-based_2021, capobianco_deep_2021}. However, all those schemes face a common limitation: they rely on a route-based representation of maritime traffic, which is viable only when the traffic is highly organized and structured. In real life, a significant fraction of AIS trajectories cannot be assigned to predefined routes \cite{nguyen_multi-task_2018, nguyen_geotracknet-maritime_2021}, limiting the practical application of clustering-based techniques in operational systems.}

In this paper, we present a novel model for \rvv{AIS-based vessel trajectory} prediction, referred to as \textit{TrAISformer}. To tackle the complexity and multimodality of \rvv{AIS vessel trajectories}, we propose a new data representation and harness the modeling capabilities of deep learning, specifically transformer architectures \cite{vaswani_attention_2017}.
Contrary to clustering-based models which constrain the trajectories to a maritime traffic graph structure \cite{pallotta_vessel_2013, mazzarella_knowledge-based_2015, forti_prediction_2020, capobianco_deep_2021, murray_ais-based_2021}, \textit{TrAISformer} is applicable to any trajectory within the region of interest, without imposing constraints on an explicit graph of maritime routes. Additionally, we re-frame the prediction as a classification-based learning problem to best forecast the positions of maritime vessels several hours into the future.

\section{Proposed Approach}
\label{sec:proposedApproach}

In this section, we detail the proposed approach. We introduce a new representation of AIS data, derive a new loss function, and provide a brief introduction of the transformer architecture used in \textit{TrAISformer}.  

\subsection{Discrete and sparse representation of AIS data}
\label{subsec:4-hot}

One of the primary challenges in trajectory prediction in general, and \rvv{AIS-based vessel trajectory} prediction in particular, is the modeling of the heterogeneous and multimodal nature of motion data given relatively low-dimensional observations. Here, we introduce a novel representation of AIS data, which addresses the heterogeneity aspect. The multimodality part will be addressed in the next subsection with a classification-based training loss.

\begin{figure}
  \centering
  \includegraphics[width=0.8\linewidth]{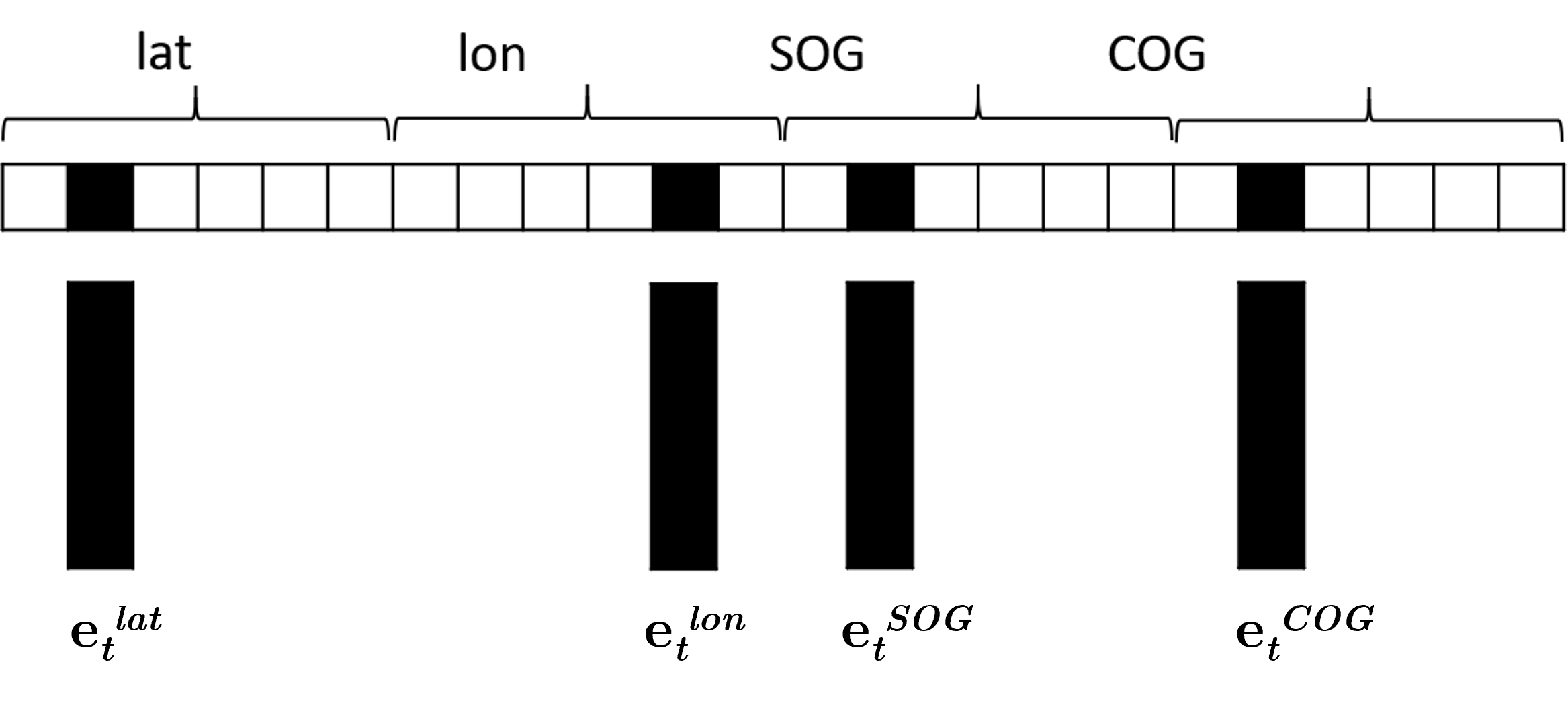}
  \centering
  \caption{{\bf Proposed representation of AIS data:} To overcome the challenge of representing the heterogeneous and multimodal nature of motion data with relatively low-dimensional observations in \rvv{AIS-based vessel trajectory} prediction, a new representation of AIS data is proposed in this study. For each attribute $att \in  \{lat, lon, SOG, COG\}$), the observed value (which is continuous) is discretized into a one-hot vector $\vect{h}^{att}_t$. Each $\vect{h}^{att}_t$ is then associated with a high dimensional real-valued embedding vector $\vect{e}^{att}_t$. } \label{fig:e}
\end{figure}

The most prevalent way to represent an AIS message is to use a 4-dimensional real-valued vector composed of the position and the velocity of the vessel, as in \eqref{eq:4d_representation}. 

However, as discussed in \cite{nguyen_multi-task_2018} and \cite{nguyen_geotracknet-maritime_2021}, it is challenging to encode complex vessel movement patterns in this feature space. A natural approach is to expand the feature space to a higher dimensional one. Specifically, instead of modeling the conditional probability distribution of the future trajectory given the past $p(\vect{x}_{T+1:T+L}|\vect{x}_{0:T})$, we consider $p(\vect{e}_{T+1:T+L}|\vect{e}_{0:T})$, 
where $\vect{e}_t \in \mathbb{R}^{d_e}$ \rvv{represents} a high-dimensional embedding vector of $\vect{x}_t$. Recently, variational autoencoders have been very successful in learning such effective mappings that encode $\vect{e}_t$ from $\vect{x}_t$, and decode $\vect{x}_t$ from $\vect{e}_t$ \cite{lecun_deep_2015, goodfellow_deep_2016, kingma_auto-encoding_2013, rezende_variational_2015, pu_variational_2016, vahdat_nvae_2020}. However, when the dimension of $\vect{e}_t$ is much higher than that of $\vect{x}_t$, the training becomes extremely difficult and prone to overfitting.

To overcome the overfitting problem, we exploit the ``four-hot'' representation vector $\vect{h}_t$, defined in our previous works \cite{nguyen_multi-task_2018},  \cite{nguyen_geotracknet-maritime_2021}. Specifically, we discretize the latitude, the longitude, the SOG, and the COG into $N_{lat}$, $N_{lon}$, $N_{SOG}$, and $N_{COG}$ bins, respectively. We then build a one-hot vector for each attribute $\{lat, lon, SOG, COG\}$. The ``four-hot'' vector $\vect{h}_t$ is the concatenation of the four one-hot vectors.
\begin{equation}
    \vect{h}_t \triangleq [1^{lat}_t, 1^{lon}_t, 1^{SOG}_t, 1^{COG}_t]^T
\end{equation}
with $1^{att}_t$ ($att$ $\in$ $\{lat, lon, SOG, COG\}$) \rvv{being} the one-hot vector of $att$. \rvv{The details are presented in Algorithm \ref{algo:fourhot}.}

Each attribute bin of $\vect{h}_t$ will be associated with a high dimensional embedding vector, denoted as $\vect{e}^{att}_t$. The embedding vector $\vect{e}_t$ of an AIS observation $\vect{x}_t$ is the concatenation of $\vect{e}^{att}_t$. This mapping is illustrated in Fig. \ref{fig:e}. The proposed approach ensures that in the embedding space, only $N_{lat} \times N_{lon} \times  N_{SOG} \times  N_{COG}$ values of $\vect{e}_t$ will be used.  By imposing this sparsity constraint, we effectively regularize the mapping and avoid overfitting when augmenting the original 4-dimensional AIS observation $\vect{x}_t$ to a much higher dimensional space of $\vect{e}_t$ \cite{ng_sparse_2011}. An \rvv{AIS vessel trajectory} is then represented by $\vect{e}_{0:T} \triangleq \{\vect{e}_0, \vect{e}_1,...,\vect{e}_T\}$.

Note that the mapping $\vect{h}_t \rightarrow \vect{e}_t$ is one-to-one, and obtaining $\vect{h}_t$ from $\vect{x}_t$ is a straightforward process. However, in the reverse direction, it is not possible find the exact $\vect{x}_t$ from $\vect{h}_t$ because of the discretization. In this paper, we employ a simplifying approximation where we use the mid-points of the bins specified by $\vect{h}_t$ to estimate $\vect{x}_t$. This approximation introduces an error equivalent to half of the resolution of the $\vect{h}_t$ bins in the estimation of $\vect{x}_t$, even when the bins estimation is perfect. Nevertheless, we argue that this inherent error is negligible for medium-range vessel trajectory prediction applications such as search and rescue, traffic control, path planning, and port congestion avoidance. To provide an illustration, let's consider a resolution of 0.01\degree \hspace{0.1em}  for $lat$ and $lon$. The approximation introduces an error of around 0.15 nautical miles (nmi). This level of error does not significantly impact the aforementioned applications, as it remains well within acceptable limits.

\subsection{Transformer architecture}
\label{subsec:transformer}

As depicted in Fig. \ref{fig:maritmeTrafficGraph}, in order 
to forecast the trajectory of a vessel correctly, a prediction model needs to capture possible long-term dependencies in the historical AIS observations. In this regard, transformer neural networks \cite{vaswani_attention_2017} naturally arise as highly suitable candidates. In this work, we adopt a transformer architecture akin to the GPT models \cite{radford_improving_2018}. The model's architecture is briefly presented in the following paragraphs. Interested readers are encouraged to refer to \cite{phuong_formal_2022} for a more complete and formal introduction to transformer.

The transformer network in \textit{TrAISformer} consists of a series of attention layers that are stacked together. Each layer functions as an auto-regressive model that employs the dot-product multiple-head self-attention mechanism:
\begin{equation}
    Attention(Q, K, V) = \text{softmax}\left (\frac{QK^{T}}{\sqrt{d_e}}\right )V,
\end{equation}
where $Q, K, V$ are linear projections of the input sequence (which is $\{\vect{e}_t\}$ for the first layer, or the output sequence of the previous layer for other layers), \rvv{and $d_e$ is the \textit{model size}, \ie the dimension of $\vect{e}_t$}. At each layer, the input sequence is projected into a new space $V$, and the output of the attention block is a weighted sum in $V$, where the weights signify the relative contribution of each time step. \rvv{These weights are computed as the $\text{softmax}$ on the dot product of $Q$ and $K$, normalized by $\sqrt{d_e}$}. The projection operators of $Q, K, V$ are learned during the training phase, and the calculation is performed in parallel. The parallel processing capability allows the model to directly retrieve information from multiple past time steps simultaneously. This is a critical advantage compared to recurrent networks, where the model has to process data sequentially and may not be able to retrieve long-term information. 

\rvv{The output of the transformer's final layer is a vector $\vect{l}_t$ with the same dimension as $\vect{h}_t$. We will present in the next section how \textit{TrAISformer} uses this output to model $p(\vect{h}_{T+l}|\vect{e}_{0:T+l-1})$.}

\subsection{Learning scheme}
\label{subsec:proposedApproach}

\begin{figure}
  \centering
  \includegraphics[width=0.8\linewidth]{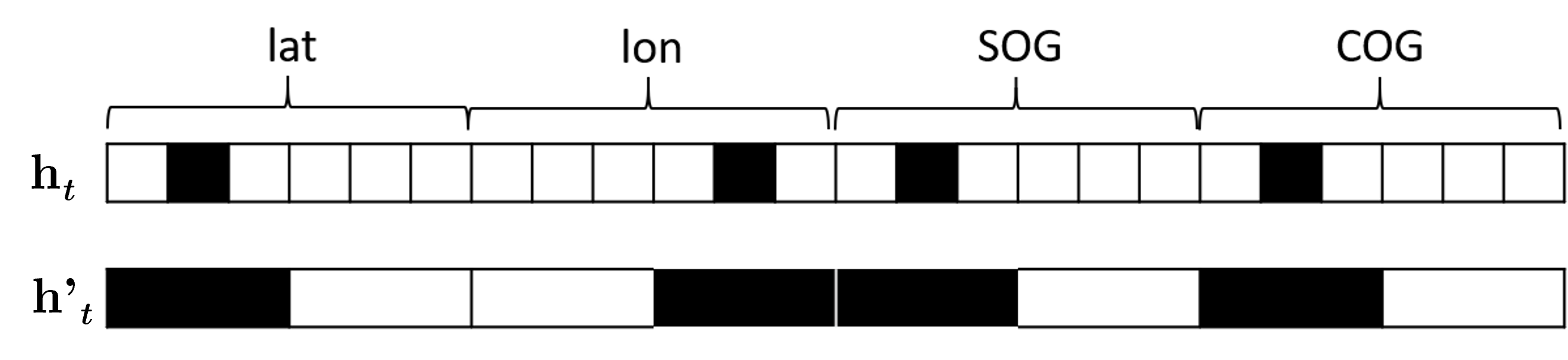}
  \centering
  \caption{{\bf Example of multi-resolution ``four-hot'' vectors for AIS data:} The model uses the fine-resolution vector $\vect{h}_t$ in the embedding module (see Fig.\ref{fig:e}), while the loss function uses both $\vect{h}_t$ and a coarse-resolution $\vect{h}'_t$.} \label{fig:multireso}
\end{figure}

\begin{figure}
  \centering
  \includegraphics[width=0.9\linewidth]{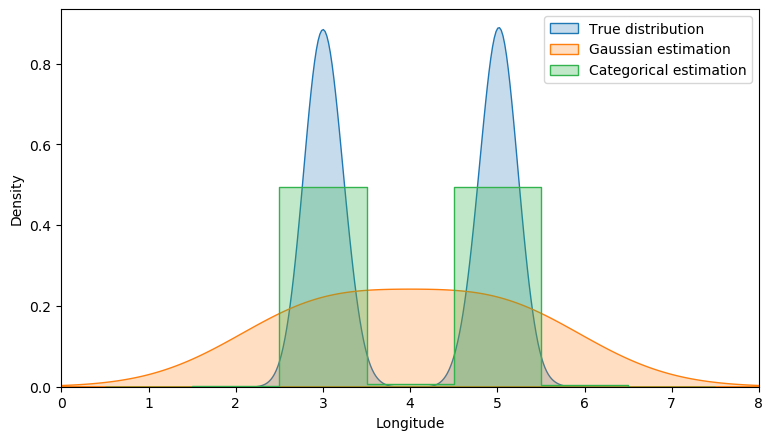}
  \centering
  \caption{{\bf Illustration of the loss function $\mathcal{L}_{CE}$ to account for multimodal posterior}: Let's consider a scenario where at a specific \textit{waypoint}, half of the vessels in the training set turn left and the other half turn right. The true distribution of the longitude at the next time step forms a bimodal normal distribution, depicted by the blue curve. If we use a real-valued scalar to represent the longitude and use $\mathcal{L}_{MSE}$ as the loss function, the implicit distribution of the model is an unimodal Gaussian distribution.  Consequently, the model tends to merge the two modes of the true distribution, as illustrated by the orange curve. By contrast, if we use a one-hot vector to represent the longitude and use $\mathcal{L}_{CE}$ as the loss function, the implicit distribution of the model is a categorical distribution. With this distribution, the model preserves the two modes, as shown by the green curve.} \label{fig:multimodal}
\end{figure}

\rvv{In the learning literature, trajectory prediction is commonly formulated as a regression problem where a model aims to output the best possible continuous-valued $\vect{e}_{T+1:T+L}$ (or $\vect{x}_{T+1:T+L}$) given the input $\vect{e}_{0:T}$ (or $\vect{x}_{0:T}$) \cite{forti_prediction_2020, capobianco_deep_2021}.} Within a deterministic setting,  the most common loss function is the mean square error, which measures the squared difference between the predicted and the actual values\cite{forti_prediction_2020, murray_ais-based_2021, dijt_trajectory_2020}:
\begin{equation}
    \mathcal{L}_{MSE} = \frac{1}{L} \sum_{l=1}^{L}||\vect{x}^{pred}_{T+l} - \vect{x}^{true}_{T+l} ||^2_2,
    \label{eq:loss_mse}
\end{equation}
where $||.||_2$ denotes the Euclidean norm $L_2$. However, this $L_2$ loss function (which can be interpreted \wrt Gaussian assumption on the conditional likelihood $p(\vect{e}_{T+l}|\vect{e}_{0:T+l-1})$ or $p(\vect{x}_{T+l}|\vect{x}_{0:T+l-1})$) may not be appropriate for posterior distributions that exhibit multimodality, as illustrated by vessels' trajectories in Fig.\ref{fig:maritmeTrafficGraph}. 
To explicitly account for multimodal posteriors, \rvv{we propose a classification-based formulation that involves modeling $p(\vect{e}_{T:T+l}|\vect{e}_{0:T+l-1})$ as a concatenation of four categorical distributions, each corresponding to an attribute $\{lat, lon, SOG, COG\}$}. Specifically, because the mapping $\vect{h}_t \rightarrow \vect{e}_t$ is one-to-one, we have:
\begin{equation}
    p(\vect{h}_{T+l}|\vect{e}_{0:T+l-1}) = p(\vect{e}_{T+l}|\vect{e}_{0:T+l-1}).
    \label{eq:obj_e2h}
\end{equation}
As $\vect{h}_t$ is a ``four-hot'' vector, \rvv{this transforms the prediction into a classification problem with four heads, each corresponding to one component of the one-hot vector $\vect{h}_t$)}. Let us denote $p_{T+l} \triangleq p(\vect{h}_{T+l}|\vect{e}_{0:T+l-1})=p(\vect{e}_{T+l}|\vect{e}_{0:T+l-1})$, the loss function is defined as:
\begin{equation}
    \mathcal{L}_{CE} \triangleq \sum_{l=1}^{L}CE(\vect{h}_{T+l}, p_{T+l}),
\end{equation}
with $CE$ \rvv{being} the cross-entropy function. \rvv{The details are presented in Algorithm \ref{algo:ce}}. We demonstrate how the proposed loss function helps maintain the multimodal characteristics of the data in Fig. \ref{fig:multimodal}. 

Note that $\vect{h}_t$ is a discrete representation of the continuous $\vect{x}_t$. This discretization can be performed at different resolutions. We empirically observed that the prediction could be marginally improved if we use a multi-resolution version of $\mathcal{L}_{CE}$ as follows (see Fig. \ref{fig:multireso}) :
\begin{equation}
    \mathcal{L}_{CE} = \sum_{l=1}^{L}CE(p_{T+l},\vect{h}_{T+l}) + \beta CE(p^{\prime}_{T+l},\vect{h}'_{T+l}).
\end{equation}
where $p'_{T+l} \triangleq p(\vect{h}'_{T+l}|\vect{e}_{0:T+l-1})$, $\vect{h}'_{T+l}$ is a coarser version of $\vect{h}_{T+l}$, $\beta$ is a scalar balancing the relative importance of the coarse-resolution loss. 

\rvv{The training procedure's specifics are outlined in Algorithm \ref{algo:train}. For the sake of notation simplicity, we present the training on a per-series basis. In practice, the model processes the data in batches.}

The proposed model is applied recursively. To predict a vessel position at time $T+l$, we sample a ``four-hot'' vector $\vect{h}^{pred}_{T+l}$ from $p(\vect{h}_{T+l}|\vect{e}_{0:T+l-1})$:
\begin{equation}
    \vect{h}^{pred}_{T+l} \sim p(\vect{h}_{T+l}|\vect{e}_{0:T+l-1})
    \label{eq:sample_h}
\end{equation}
and compute the ``pseudo-inverse'' of the sampled "four-hot" vector to output the new position $\vect{x}^{pred}_{T+l}$. 
The latter is subsequently fed into the network to sample similarly a position at the next time step. We repeat this sampling procedure until we achieve the desired trajectory length.  Multiple runs of this sampling procedure can be performed for a given \rvv{AIS vessel trajectory} to generate different possible predicted paths. This stochastic procedure allows us to address the fact that two vessels currently having similar movement patterns at present may diverge in their trajectories at the next waypoint. \rvv{The details are presented in Algorithm \ref{algo:predict}}. We demonstrate in Section \ref{sec:experiments} that if we do not sample $\vect{h}^{pred}_{T+l}$ from $p(\vect{h}_{T+l}|\vect{e}_{0:T+l-1})$ according to \eqref{eq:sample_h}, the performance of the model will degrade.

\begin{figure}
  \centering
  \includegraphics[width=\linewidth]{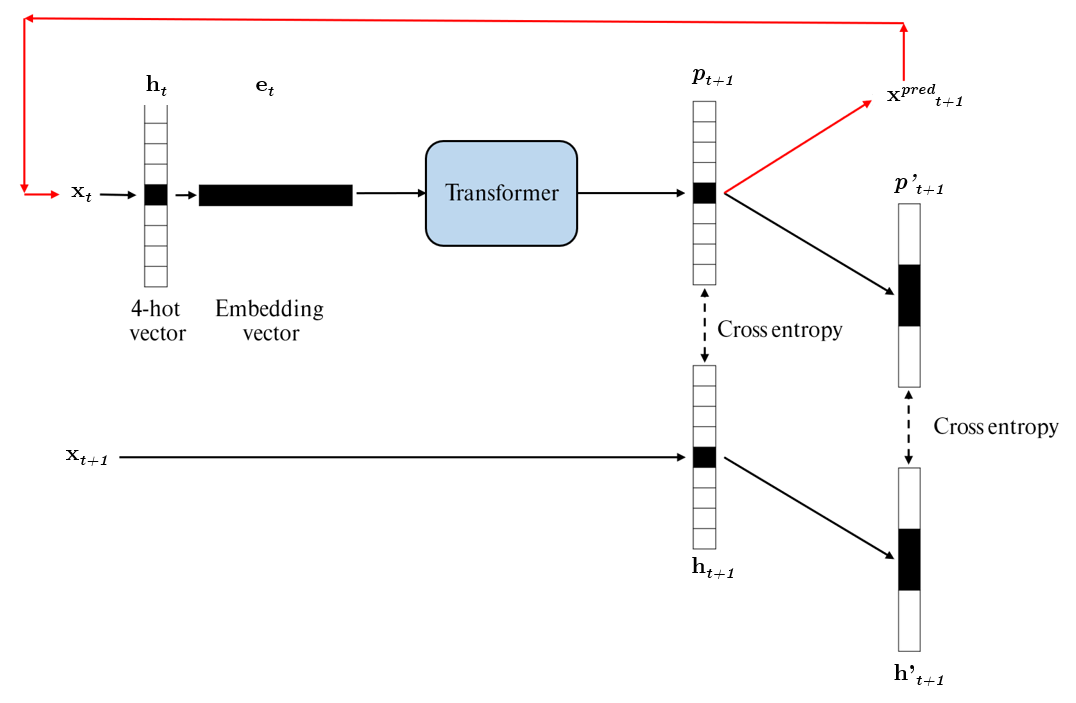}
  \centering
  \caption{{\bf Sketch of the \textit{\textbf{TrAISformer}} architecture:} each AIS observation $\vect{x}_t$ is discretized into a ``four-hot'' vector $\vect{h}_t$ (for visualization purposes, we illustrate a one-hot vector instead of a ``four-hot'' vector for $\vect{h}_t$). Subsequently, each $\vect{h}_t$ is paired with a high dimensional real-valued embedding vector $\vect{e}_t$. The sequence of embeddings the $\vect{e}_{0:t}$ will be fed into a transformer network to predict $p_{t+1} \triangleq p(\vect{h}_{t+1}|\vect{e}_{0:t})$.  During the training phase, the model is optimized to minimize the cross-entropy loss between the true value $\vect{h}_{t+1}$ and $p_{t+1}$. To enhance prediction accuracy, we introduce a ``multi-resolution'' loss. This involves calculating the cross-entropy at different spatial resolutions of $\vect{h}_{t+1}$. In the forecasting phase,  we generate vessel positions recursively. We sample $\vect{h}^{pred}_{t+1}$  from  $p_{t+1}$, calculate the ``pseudo-inverse'' of it to derive $\vect{x}^{pred}_{t+1}$. The predicted $\vect{x}^{pred}_{t+1}$ is fed back into the network to sample the next position (as shown by the red path in the diagram). This iterative process continues until we reach the desired prediction horizon $L$.} \label{fig:architecture}
\end{figure}

A sketch of the resulting \textit{TrAISformer} architecture is shown in Fig. \ref{fig:architecture}. 

\rvv{\begin{algorithm}[h]
\caption{fourhot($\vect{x}_t, \vect{R}, SOG_{\texttt{max}}, \vect{N}$). \label{algo:fourhot}}
\DontPrintSemicolon
\KwDescription{Create "four-hot" vector.}
\KwIn{AIS observation \newline
$\vect{x}_t \triangleq \left[lat, lon, SOG, COG \right]^T$, \newline
the limits of the ROI $\vect{R} \triangleq \left[lat_{\texttt{min}}, lat_{\texttt{max}}, lon_{\texttt{min}}, lon_{\texttt{max}} \right]^T$, \newline
$SOG_{\texttt{max}}$, \newline
the numbers of bins \newline $\vect{N} \triangleq [N^{lat}$, $N^{lon}$, $N^{SOG}$, $N^{COG}]^T$.}
\KwOut{``Four-hot'' vector $\vect{h}_t$.}

\tcp{{Create the one-hot vector for each attribute.}}
$1^{lat}_t = \texttt{onehot}(\vect{x}^{lat}_{t}, lat_{\texttt{min}}, lat_{\texttt{max}}, N^{lat})$

$1^{lon}_t = \texttt{onehot}(\vect{x}^{lon}_{t}, lon_{\texttt{min}}, lon_{\texttt{max}}, N^{lon})$

$1^{SOG}_t = \texttt{onehot}(\vect{x}^{SOG}_{t}, 0, ${\footnotesize $SOG$}$_{\texttt{max}}, N^{SOG})$

$1^{COG}_t = \texttt{onehot}(\vect{x}^{COG}_{t}, 0, 360, N^{COG})$

\tcp{{Concatenate the one-hot vectors}}
$\vect{h}_t = [1^{lat}_t, 1^{lon}_t, 1^{SOG}_t, 1^{COG}_t]^T$

\KwReturn{$\vect{h}_t$}
        
\end{algorithm}

\begin{algorithm}[h]
\caption{ce\_loss($\vect{h}_t$, $\vect{l}_t, \vect{N}$). \label{algo:ce}}
\DontPrintSemicolon
\KwDescription{Calculate the cross-entropy loss $\mathcal{L}_{CE}$.}
\KwIn{"four-hot" vector $\vect{h}_t$, \newline
the output of the transformer $\vect{l}_t$, \newline
the numbers of bins $\vect{N}$.}
\KwOut{the cross-entropy $CE(\vect{h}_t, \vect{l}_t)$.}

\tcp{{Split $\vect{h}_t$ back into 4 one-hot vectors, each corresponding to an attribute of the AIS observation.}}
$1^{lat}_t, 1^{lat}_t, 1^{SOG}_t, 1^{COG}_t$ = $split(\vect{h}_{t}, \vect{N})$

\tcp{{Split $\vect{l}_t$ into 4 heads.}}
$\vect{l}^{lat}_{t}, \vect{l}^{lon}_{t}, \vect{l}^{SOG}_{t}, \vect{l}^{COG}_{t}$ = $split(\vect{l}_{t}, \vect{N})$

\tcp{{Calculate the cross-entropy for each head.}}

$p^{lat}_t = CE(\texttt{Categorical}(logit=\vect{l}^{lat}_t), 1^{lat}_t)$

$p^{lon}_t = CE(\texttt{Categorical}(logit=\vect{l}^{lon}_t), 1^{lon}_t)$

\mbox{$p^{{\footnotesize SOG}}_t = CE(\texttt{Categorical}(logit=\vect{l}^{{\footnotesize SOG}}_t), 1^{{\footnotesize SOG}}_t)$}

\mbox{$p^{{\footnotesize COG}}_t = CE(\texttt{Categorical}(logit=\vect{l}^{{\footnotesize COG}}_t),1^{{\footnotesize COG}}_t)$}

\tcp{{Calculate ``total'' cross-entropy.}}
$ce\_with\_logit(\vect{h}_t, \vect{l}_t) = p^{lat}_{t}*p^{lon}_{t}*p^{SOG}_{t}*p^{COG}_{t}$ 

\KwReturn{$ce\_with\_logit(\vect{h}_t, \vect{l}_t)$}
        
\end{algorithm}

\begin{algorithm}[h]
\caption{trAISformer\_train($\{\vect{x}_{0:T+L}\}$, $\Theta$, $\vect{R}$, $SOG_{\texttt{max}}$, $\vect{N}$, $\beta$).\label{algo:train}}
\DontPrintSemicolon
\KwDescription{Train \textit{TrAISformer}.}
\KwIn{The training set $\{\vect{x}_{0:T+L}\}$, \newline 
the set of \textit{TrAISformer}'s parameters $\Theta$,  \newline
the limits of the ROI $\vect{R}$, \newline
$SOG_{\texttt{max}}$, \newline
the numbers of bins $\vect{N}$, \newline
coefficient $\beta$.
}
\KwOut{The learned set of parameters $\Theta$.}

\For{$\vect{x}_{0:T+L}$ in $\{\vect{x}_{0:T+L}\}$}   
{
    \tcp{{Create the "four-hot" vectors at different resolutions.}}
    \For{$t$ in $0:T+L$}
    {
        $\vect{h}_t = \texttt{fourhot}(\vect{x}_t, \vect{R}, SOG_{\texttt{max}}, \vect{N}$)

        $\vect{h}'_t = \texttt{fourhot}(\vect{x}_t, \vect{R}, SOG_{\texttt{max}}, \vect{N}/3$)
    }
    \tcp{{Get the embeddings and apply the transformer.}}
    $\vect{e}_{0:T+L-1} = \texttt{embedding}(\vect{h}_{0:T+L-1})$

    $\vect{l}_{1:T+L} = \texttt{transformer}(\vect{e}_{0:T+L-1})$

    $\vect{l}'_{1:T+L} = \texttt{conv1d}(\vect{l}_{1:T+L}, kernel=[1/3, 1/3, 1/3]^T, stride=3)$

    \tcp{{Calculate the loss.}}
    $\mathcal{L}_{CE} = 0$
    
    \For{$l$ in $1:L$}
    {
        $ce(\vect{h}_t, \vect{l}_t) = \texttt{ce\_loss}(\vect{h}_t, \vect{l}_t, \vect{N})$

        $ce(\vect{h}'_t, \vect{l}'_t) = \texttt{ce\_loss}(\vect{h}'_t, \vect{l}'_t, \vect{N}/3)$

        $\mathcal{L}_{CE} = \mathcal{L}_{CE} + ce(\vect{h}_t, \vect{l}_t) + \beta*ce(\vect{h}'_t, \vect{l}'_t)$
    }

    \tcp{{Optimize $\Theta$.}}
    $\Theta = \texttt{AdamW}(\mathcal{L}_{CE}, \Theta, \vect{x}_{0:T+L})$ 
}
\KwReturn{$\Theta$}
\end{algorithm}

\begin{algorithm}[h]
\caption{trAISformer\_predict($\vect{x}_{0:T}$, $\Theta$, $\vect{R}$, $SOG_{\texttt{max}}$, $\vect{N}$, $L$).\label{algo:predict}}
\DontPrintSemicolon
\KwDescription{Use \textit{TrAISformer} to predict vessel trajectory.}
\KwIn{The initial segment of the trajectory to predict $\vect{x}_{0:T}$, \newline 
the trained \textit{TrAISformer}'s parameters $\Theta$,  \newline
the limits of the ROI $\vect{R}$, \newline
$SOG_{\texttt{max}}$, \newline
the numbers of bins $\vect{N}$, \newline
the prediction horizon $L$.
}
\KwOut{The predicted trajectory $\vect{x}_{0:T+L}$.}

\tcp{{Create the "four-hot" vectors of the initial segment.}}
\For{$t$ in $0:T$}
{
    $\vect{h}_t = \texttt{fourhot}(\vect{x}_t, \vect{R}, SOG_{\texttt{max}}, \vect{N}$)
}

\tcp{{Get the embeddings of the initial segment.}}
$\vect{e}_{0:T} = \texttt{embedding}(\vect{h}_{0:T})$

\tcp{Iterate over the prediction horizon.}
\For{$l$ in $1:L$}
{
    $\vect{l}_{1:T+l} = \texttt{transformer}(\vect{e}_{0:T+l-1})$

    \tcp{{Split $\vect{l}_{T+l}$ into 4 heads.}}
    $\vect{l}^{lat}_{T+l}, \vect{l}^{lon}_{T+l}, \vect{l}^{SOG}_{T+l}, \vect{l}^{COG}_{T+l}$ = $split(\vect{l}_{T+l}, \vect{N})$

    \tcp{{Create the categorical distributions and sample $\vect{h}^{att}_{T+l}$ from them.}}
    \For{$att$ in $\{lat, lon, SOG, COG \}$ }
    {
        $\vect{h}^{att}_{T+l} \sim \texttt{Categorical}(logit=\vect{l}^{att}_{T+l})$
    }

    \tcp{Get the predicted $\vect{h}_{T+l}$, $\vect{e}_{T+l}$, and $\vect{x}_{T+l}$}
    
    $\vect{h}_{T+l} = [\vect{h}^{lat}_{T+l}, \vect{h}^{lon}_{T+l}, \vect{h}^{SOG}_{T+l}, \vect{h}^{COG}_{T+l}]^T$
    
    $\vect{e}_{T+l} = \texttt{embedding}(\vect{h}_{T+l})$

    $\vect{x}_{T+l} = \texttt{pseudo-inverse}(\vect{h}_{T+l})$

}

\KwReturn{ $\vect{x}_{0:T+L}$}
\end{algorithm}
}
\section{Experiments and results}
\label{sec:experiments}

In this section, we report the experimental results of our model on a real AIS dataset introduced in \cite{capobianco_deep_2021}. We include a benchmarking w.r.t state-of-the-art methods for a prediction horizon up to 15 hours. Additionally, we also present an ablation study to assess the relevance of each component of the proposed model. To facilitate the reproducibility of the work in this paper, we chose a  publicly available AIS dataset and made the code of the model available at https://github.com/CIA-Oceanix/TrAISformer.

\subsection{Experimental set-up}
\textbf{Dataset}: We tested \textit{TrAISformer} on a public AIS dataset provided by the Danish Maritime Authority (DMA)\footnote{https://dma.dk/safety-at-sea/navigational-information/ais-data}. The dataset comprises AIS observations of cargo and tanker vessels from January 01, 2019 to March 31, 2019. The Region of Interest (ROI) is a rectangle from (55.5\degree, 10.3\degree) to (58.0\degree, 13.0\degree). Prior to preprocessing, the raw dataset contained approximately 712 million AIS messages. 
We used AIS data from January 01, 2019 to March 10, 2019 and from March 11, 2019 to March 20, 2019 to train the model and tune the hyper-parameters, respectively. The test set comprises AIS data from March 21, 2019 to March 31, 2019. A subset of this dataset was exploited in \cite{capobianco_deep_2021} to evaluate state-of-the-art models for \rvv{AIS-based vessel} trajectory prediction, including deep learning models.

\textbf{Data pre-processing}: \rvv{AIS data often contain outliers and missing data, which can pose challenges to the prediction.  
In the training phase, the presence of outliers and missing data introduces additional noise and uncertainty, potentially affecting the convergence of the learning process. During the evaluation phase, missing data prevents us from calculating the prediction errors, while outliers can lead to an inaccurate assessment of prediction accuracy. To mitigate the impact of outliers and missing data, we implemented the following preprocessing steps:}
\begin{itemize}
    \item Remove AIS messages with unrealistic speed values (SOG $\ge$ 30 knots);
    \item Remove moored or at-anchor vessels;
    \item Remove AIS observations within 1 nautical mile distance to the coastline;
    \item Split non-contiguous voyages into contiguous ones. A contiguous voyage \cite{nguyen_multi-task_2018, nguyen_geotracknet-maritime_2021} is a voyage whose the maximum interval between two consecutive AIS messages is smaller than a predefined value, here 2 hours;
    \item Remove AIS voyages whose length is smaller than 20 or those that last less than 4h;
    \item Remove abnormal messages. An AIS message is considered abnormal if the empirical speed (calculated by dividing the distance traveled by the corresponding interval between the two consecutive messages) is unrealistic, here above 40 knots;
    \item Down-sample AIS data with a sampling rate of 10-minute;
    \item Split long voyages into shorter ones with a maximum sequence length of 20 hours.
\end{itemize}

\textbf{Hyper-parameters}: the results reported in this paper were obtained using a transformer architecture with 8 layers. Each layer contains 8 attention heads. The resolution of the ``four-hot'' vector $\vect{h}_t$ was set to 0.01\text{\degree}  for $lat$ and $lon$, 1 knot for $SOG$ and 5\text{\degree} for $COG$. With this resolution, the corresponding sizes of $\vect{e}_t^{lat}, \vect{e}_t^{lon}, \vect{e}_t^{SOG}, \vect{e}_t^{COG}$ were $256, 256, 128$ and $128$ for the ROI reported in this paper. This resulted in a $768$-dimensional embedding $\vect{e}_t$. The coarse vector $\vect{h}'_t$ was obtained by merging three consecutive bins of $\vect{h}_t$.
We noticed that when we reduced or increased the resolution of $\vect{h}_t$ by 2, the difference in the results was negligible. The historical sequence length $T$ was set to 3 hours and the prediction horizon $L$ was up to 15 hours. The model was trained using AdamW optimizer \cite{loshchilov_decoupled_2019} with cyclic cosine decay learning rate scheduler \cite{loshchilov_sgdr_2017}. The learning rate was set to $6e^{-4}$. Other implementation details can be found in the GitHub repository that we shared above. We trained the model on a single GTX 1080 Ti GPU over 50 epochs with early stopping. In terms of computational complexity, it took $\sim$60 minutes to process 10 days of data in the test set, which suggests that the model can run in real-time \cite{nguyen_detection_2020}. 

\textbf{Benchmark models}: \rvv{we compare the performance of \textit{TrAISformer} against different state-of-the-art deep learning models:
LSTM seq2seq \cite{forti_prediction_2020}, convolutional seq2seq \cite{gehring_convolutional_2017}, seq2seq with attention \cite{capobianco_deep_2021}, \cite{murray_ais-based_2021}, \textit{GeoTrackNet} \cite{nguyen_geotracknet-maritime_2021}.} 

It is challenging to conduct a fair quantitative comparison with clustering-based methods \cite{pallotta_vessel_2013, mazzarella_knowledge-based_2015, forti_prediction_2020, capobianco_deep_2021, murray_ais-based_2021, suo_ship_2020}. First,  those methods did not state clearly how to address clustering noise and small clusters. Second, most of them use a DBSCAN clustering, which is sensitive to hyper-parameters \cite{mazzarella_knowledge-based_2015, murray_ais-based_2021, suo_ship_2020}. Different sets of hyper-parameters could lead to very different results. Third, as mentioned in Section \ref{sec:related_work}, clustering-based approaches, such as \cite{murray_ais-based_2021}, assume vessels' trajectories belong to a predefined graph of maritime routes. This assumption does not hold for the considered dataset. This is the reason why \cite{capobianco_deep_2021} restricted their analysis to a subset of the whole dataset. That subset is composed of tankers' trajectories for a few predefined routes\footnote{We may point out that, contrary to the whole dataset, that subset has not been made available.}. Though they only involve a subset of trajectories compared with the other benchmarked approaches, we regard the resulting score in \cite{capobianco_deep_2021} as a score under a best-case scenario for clustering-based methods for the considered ROI.

As the benchmarked models are not public, we conducted an independent implementation and fine-tuned each model to get optimal outcomes.
  
\textbf{Evaluation criteria}: for each prediction, the prediction error at time step $t$ is calculated as the haversine distance between the true position and the predicted one:
\begin{equation}
    d_k = 2R\arcsin\left({\sqrt{\text{sin}^2(\bar{\phi}) + \text{cos}(\phi_1)\text{cos}(\phi_2)\text{sin}^2(\bar{\lambda})}}\right),
\end{equation}
with $R$ the radius of the Earth, $\bar{\phi} \triangleq 0.5(\phi_2-\phi_1)$, $\bar{\lambda} \triangleq 0.5(\lambda_2-\lambda_1)$, $\phi_1$ and $\phi_2$ denote the latitudes, $\lambda_1$ and $\lambda_2$ denote the longitudes of the predicted position and the true position, respectively.  

We used a \textit{best-of-N} criterion, \ie for each model, we sampled $N$ predictions for each target trajectory and reported the best result. In this paper, $N=16$. This criterion allows us to account for the effect of multimodality.


\subsection{Results}

Table \ref{tab:result} shows the average prediction errors evaluated at 1, 2, and 3 hours ahead horizons. The ROI contains several waypoints,  rendering prediction for time horizons ranging from 1 to 15 hours highly challenging. In the case of incorrect prediction of turning directions at the waypoints, the prediction errors of a model increase significantly, often above a few nautical miles (nmi), as demonstrated by the benchmarked models in Table \ref{tab:result}. 
\textit{TrAISformer} outperforms all the benchmarked models by a large margin. For instance, for the 2-hour-ahead prediction, it is the only model with an average error below one nautical mile (41\% better than the second best model \textit{GeoTracknet}). 
These results confirm the capability of \textit{TrAISformer} to capture the multimodal nature of vessel trajectories, extract pertinent long-term dependencies, and deliver accurate predictions of vessel paths.

\textit{TrAISformer} improves by a factor of 2 the performance of the model proposed in \cite{capobianco_deep_2021}, which is one of the current state-of-the-art schemes,  with respective scores of 0.94 nmi and 1.93 nmi. We may recall that the performance of this clustering-based scheme refers to a best-case scenario, as it only involves tankers' trajectories for a few maritime routes in the case-study region.
We also note that the direct application of state-of-the-art deep learning schemes on the 4-dimensional AIS feature vector, namely LSTM seq2seq \cite{forti_prediction_2020}, convolutional seq2seq \cite{gehring_convolutional_2017}, seq2seq with attention \cite{capobianco_deep_2021}, \cite{murray_ais-based_2021}, transformer \cite{vaswani_attention_2017, radford_improving_2018} (see Table. \ref{tab:ablation}) leads to poor prediction performances (mean error greater than 6 nmi for a 2-hour-ahead prediction). The second best approach is our previous work \textit{GeoTrackNet} \cite{nguyen_multi-task_2018, nguyen_geotracknet-maritime_2021}. It shares two key features with \textit{TrAISformer}: i) a similar sparse high-dimensional representation of AIS data and ii) a probabilistic neural-network-based learning scheme. However, \textit{GeoTrackNet} uses a Variational Recurrent Neural Network (VRNN) \cite{chung_recurrent_2015} instead of a transformer architecture to capture the temporal patterns in the AIS data. The improved performance of \textit{TrAISformer} over \textit{GeoTrackNet}  suggests that transformers may be a better neural architecture for AIS data than VRNN. 

To further highlight the importance of the probabilistic feature of \textit{TrAISformer}, we report the performance of a deterministic version---denoted as \textit{TrAISformer\_No-Stoch}---of \textit{TrAISformer}. This model outputs the "four-hot" vector with the highest probability, \ie $\vect{h}^{pred}_{T+l} = \argmax_{\vect{h}} p(\vect{h}_{T+l}|\vect{e}_{0:T+l-1})$, instead of sampling $\vect{h}^{pred}_{T+l}$ like in \eqref{eq:sample_h}. The decrease in the prediction performance (from 0.94 to 2.88 for the 2-hour-ahead prediction)  demonstrates the importance of a multimodal representation of vessels' trajectories for the considered case-study. As pointed out previously, two vessels departing from the same port, having the same current position and velocity, may follow different paths at the next waypoint, making it impossible for the prediction model to produce correct deterministic forecasts all the time. Models that are capable of predicting multiple possibilities are more relevant. Yet, the deterministic version of \textit{TrAISformer} is still much better than standard seq2seq models, which again stresses the relevance of the transformer architecture as well as of the considered representation of AIS feature vector.            

\begingroup
\begin{table}[]
    \caption {{\bf Mean prediction performance of the benchmarked models (in nautical miles)}.}
    \label{tab:result}
    \centering
    \begin{threeparttable}
    \begin{tabular}{l*{3}c}
    \toprule
    Model &  1h & 2h & 3h \\
    \midrule 
    LSTM\_seq2seq & 5.83 & 8.39  & 11.64 \\
    \midrule 
    Conv\_seq2seq  & 4.23 & 6.77 & 9.66\\
    \midrule 
    LSTM\_seq2seq\_att  & 3.35 & 6.41 & 9.65 \\
    \midrule 
    \textit{Clustering\_LSTM\_seq2seq\_att}\tnote{1}  \cite{capobianco_deep_2021} & \textit{0.78} & \textit{1.93} & \textit{3.66} \\  
    \midrule 
    GeoTrackNet \cite{nguyen_geotracknet-maritime_2021}  & 0.72 & 1.59 & 2.67 \\

    \midrule 
    TrAISformer & \textbf{0.48} & \textbf{0.94} & \textbf{1.64} \\
    \midrule 
    \midrule 
    \color{black}{TrAISformer\_No-Stoch} & 1.28  & 2.88 & 5.02 \\
    \bottomrule
    \end{tabular}
    \begin{tablenotes}
    \item[1] The result from \cite{capobianco_deep_2021} was evaluated on a subset of the whole dataset, which comprises only tankers' trajectories for a predefined number of maritime routes. As such, this is regarded as a best-case scenario for clustering-based models.
  \end{tablenotes}
    \end{threeparttable}
\end{table}
\endgroup

\begin{figure}
  \centering
  \includegraphics[width=\linewidth]{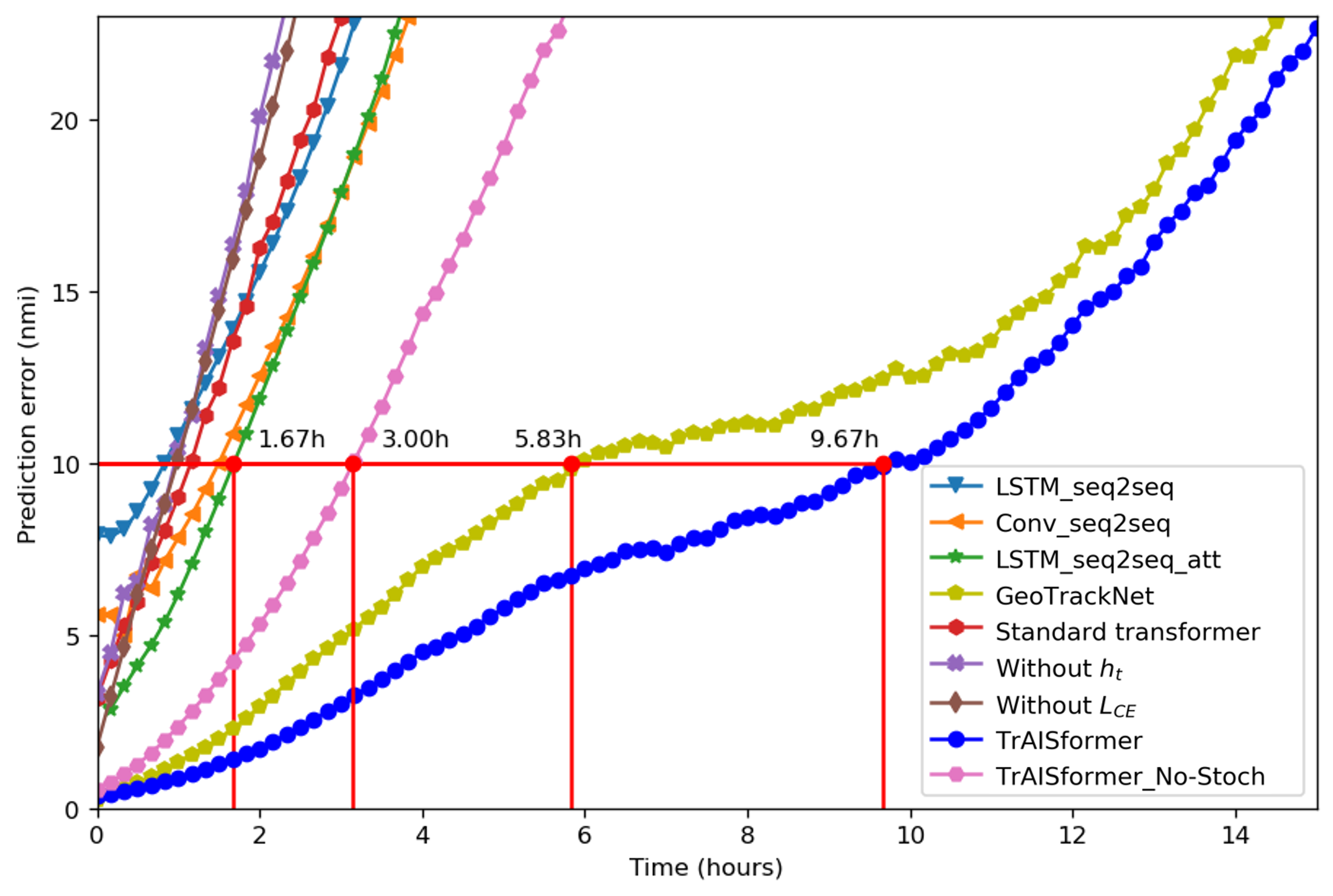}
  \centering
  \caption{{\bf Prediction performance w.r.t. prediction time horizon}: we plot for each benchmarked model the mean prediction performance for prediction time horizons from 10 minutes to 15 hours. We also highlight the time horizon up to which the performance of a given model remains below the maximum visibility under good weather conditions (i.e., 10 nmi).} \label{fig:pred_error}
\end{figure}

In maritime downstream tasks, a crucial factor to consider when using a prediction model is the maximum meaningful prediction horizon, which is the longest time horizon where a prediction is still useful. In some scenarios, such as search and rescue operations, a prediction is deemed helpful if the prediction error is smaller than the visibility, which is generally assumed to be 10 nmi under clear weather conditions \cite{blance_nories_2019}.  Fig. \ref{fig:pred_error} depicts such prediction horizons of the benchmarked models. In the best case scenario, \textit{TrAISformer} can extend the prediction horizon by a factor of $\sim$5.8 compared with current state-of-the-art methods (9.67h for \textit{TrAISformer} \vs 1.67h for LSTM\_seq2seq\_att). 

Fig. \ref{fig:predictions} displays some examples of the predictions made by \textit{TrAISformer} and the other benchmarked methods. \textit{TrAISformer} successfully samples realistic turning directions to forecast the potential paths taken by vessels. We recall here that for probabilistic models such as \textit{GeoTrackNet}, \textit{TrAISformer}, we report among 16 sampled trajectories the one closest to the real trajectory. We may highlight that the model applies not only to the main maritime routes (the first three columns) but also to less frequent ones (the last column). By contrast, clustering-based methods struggle in such cases. The four examples show the relatively poor performance of the direct application of sequence-to-sequence deep learning models. \textit{GeoTrackNet} samples realistic trajectories for the first three examples, though not as close to the real ones as the ones predicted by \textit{TrAISformer}. However, for the last example, it performs poorly, while \textit{TrAISformer} still succeeds in sampling a realistic path.

\newcommand{\cwidth}{0.24\linewidth}%
\newcommand{\twidth}{0.5\linewidth}%
\begin{figure*}
    \centering
	\begin{subfigure}[b]{0.02\linewidth}
	    \rotatebox[origin=t]{90}{\scriptsize LSTM\_seq2seq\_att}\vspace{2.7\linewidth}
	\end{subfigure}%
	\begin{subfigure}[t]{0.96\linewidth}
	    \centering
		\includegraphics[width=\cwidth,clip]{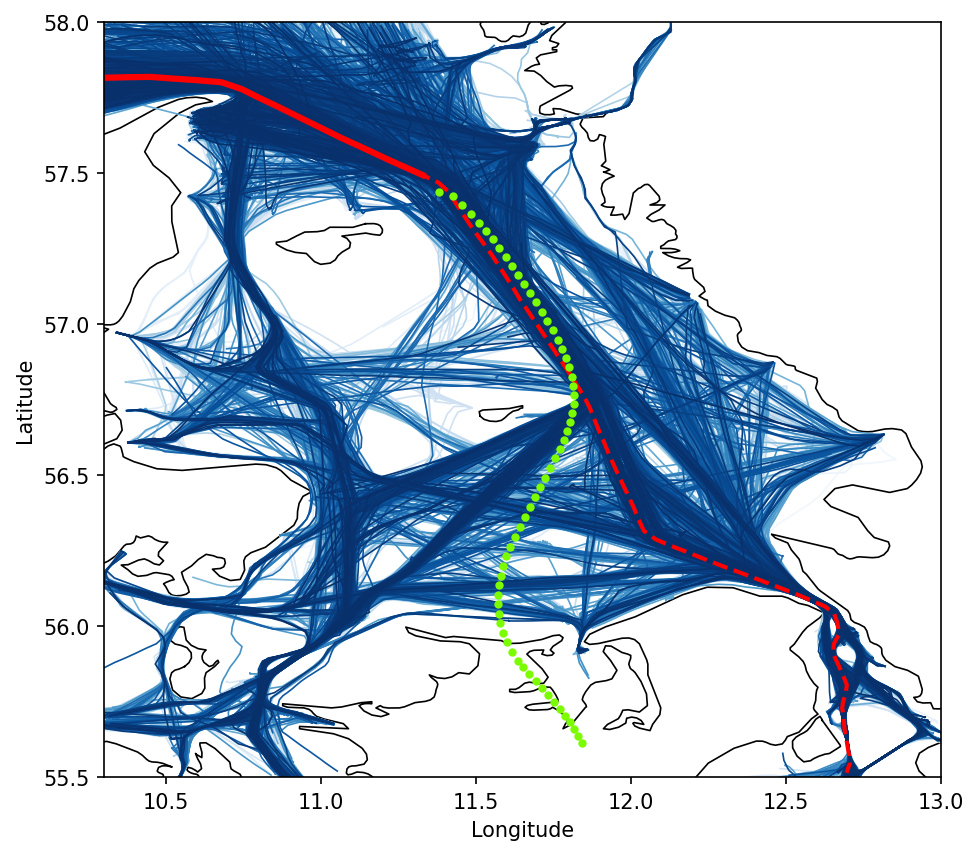}
		\includegraphics[width=\cwidth,clip]{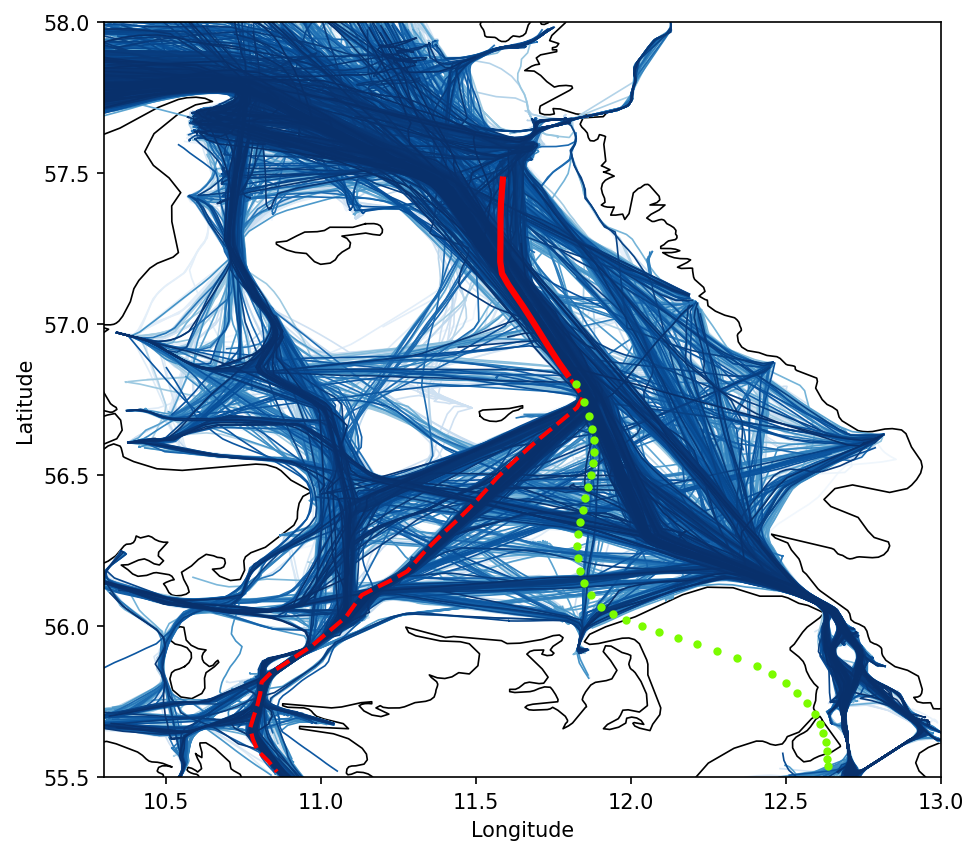}
		\includegraphics[width=\cwidth,clip]{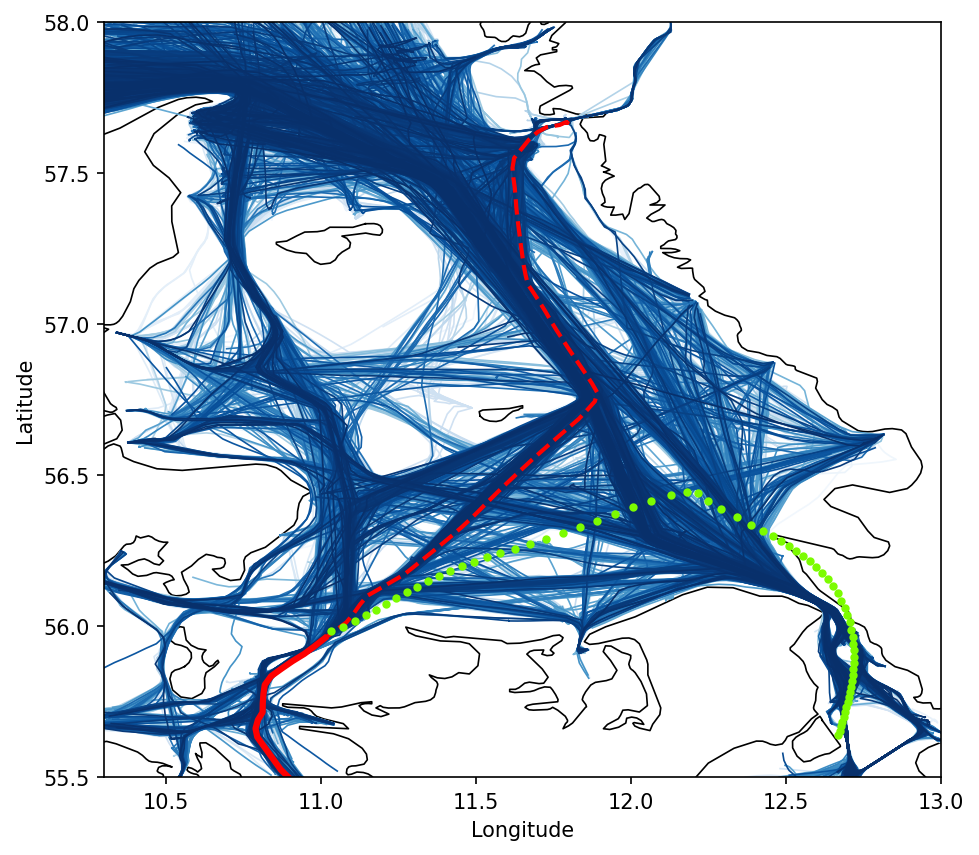}
		\includegraphics[width=\cwidth,clip]{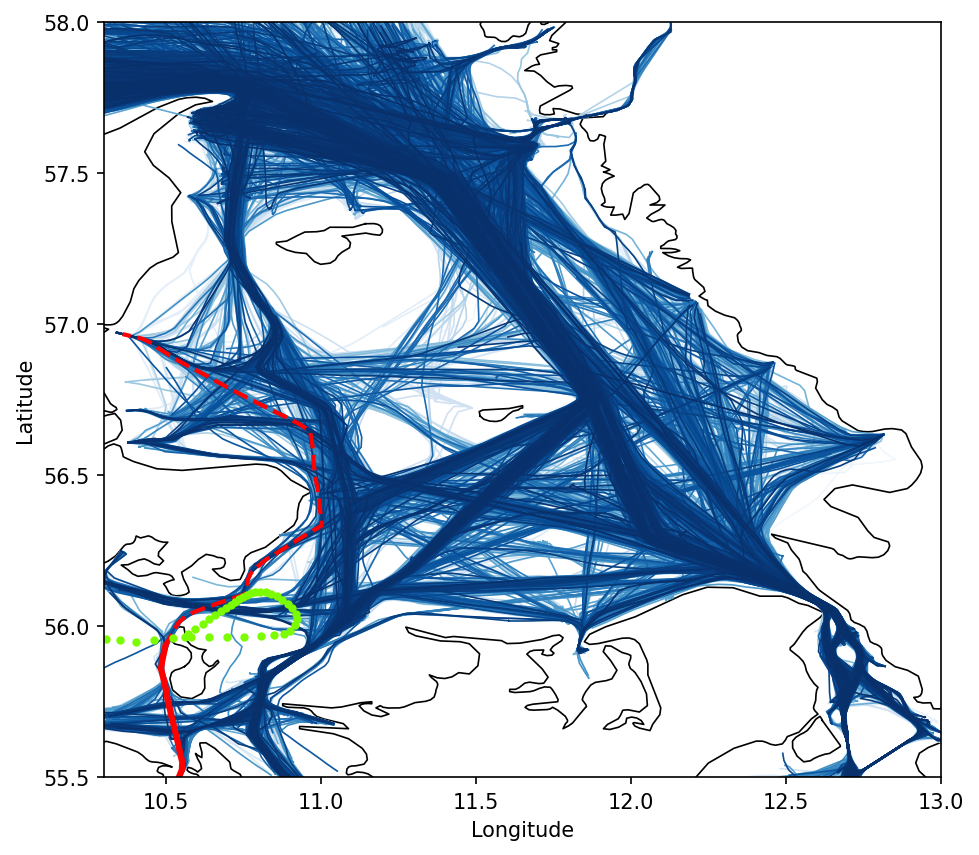}
	\end{subfigure}%

	\begin{subfigure}[b]{0.02\linewidth}
	    \rotatebox[origin=t]{90}{\scriptsize GeoTrackNet}\vspace{3.5\linewidth}
	\end{subfigure}%
	\begin{subfigure}[t]{0.96\linewidth}
	    \centering
		\includegraphics[width=\cwidth,clip]{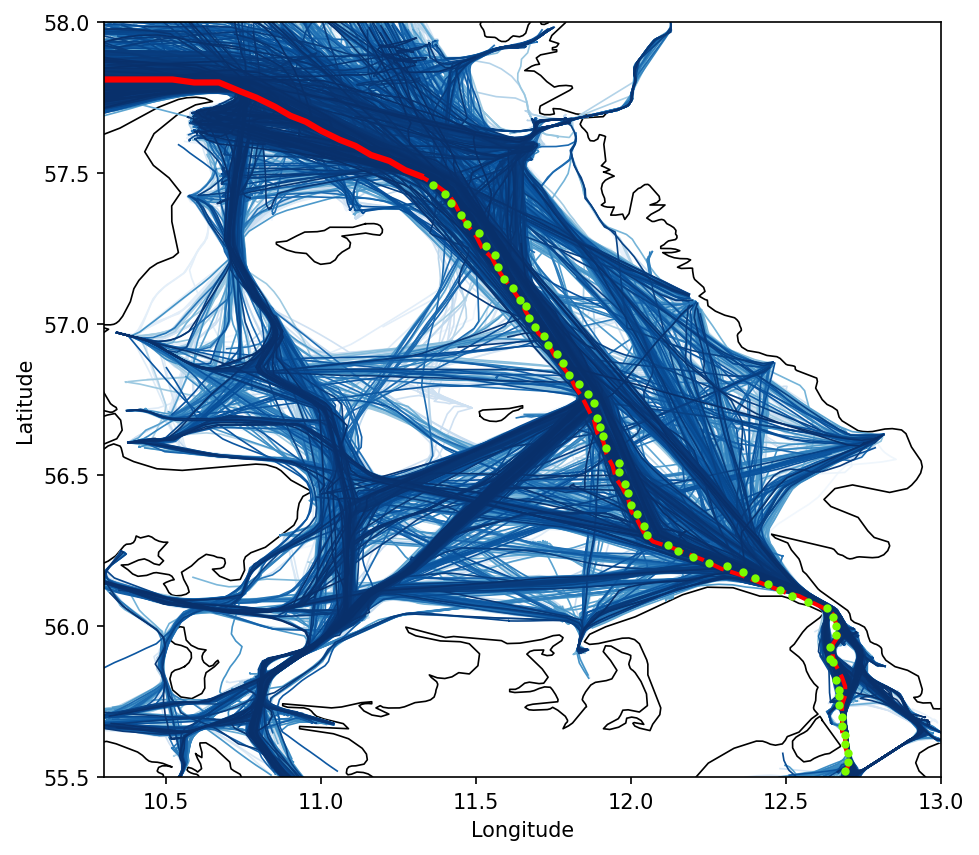}
		\includegraphics[width=\cwidth,clip]{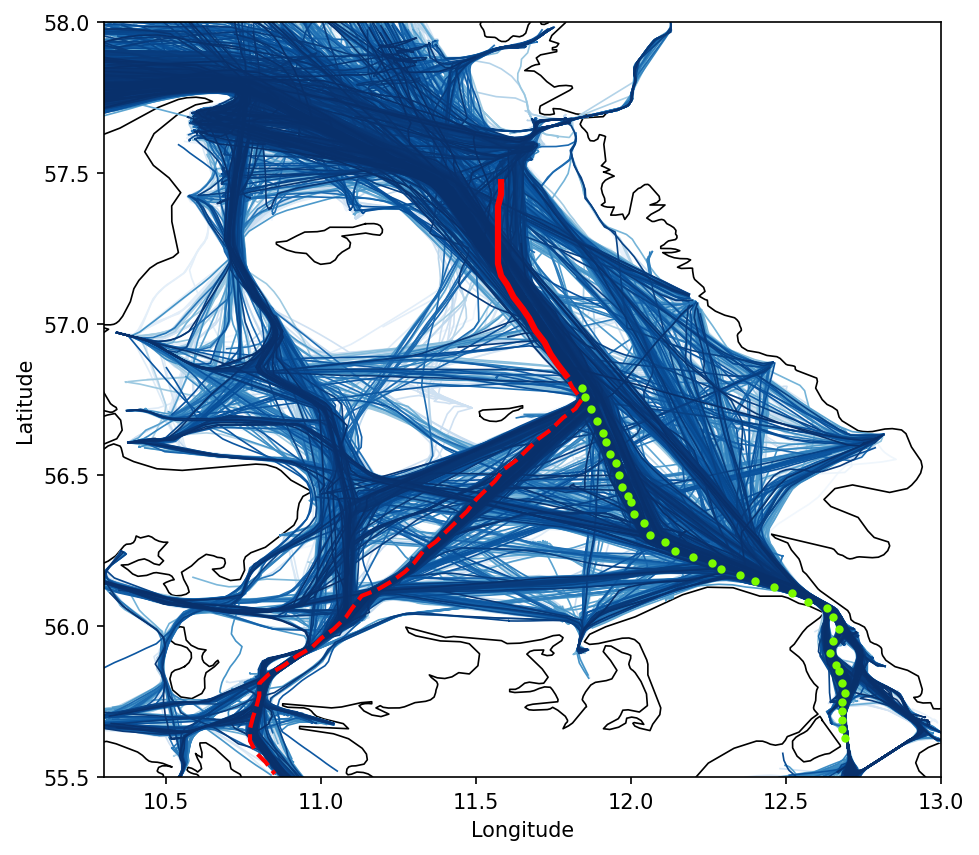}
		\includegraphics[width=\cwidth,clip]{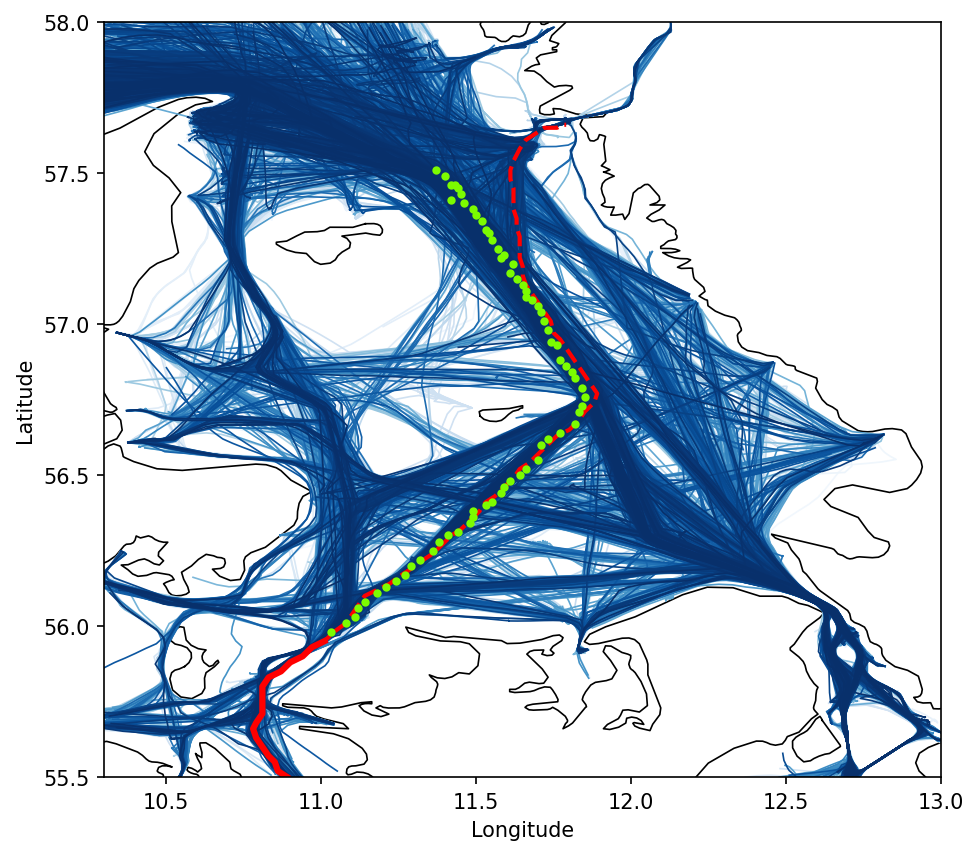}
		\includegraphics[width=\cwidth,clip]{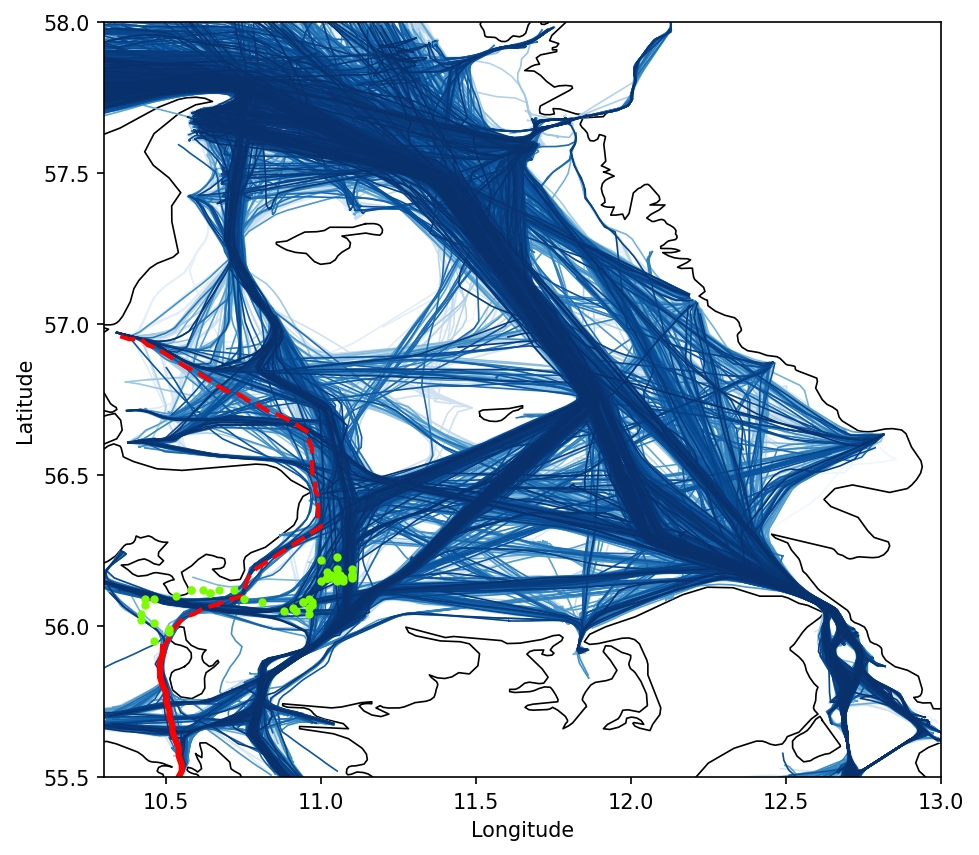}
	\end{subfigure}%
	
	\begin{subfigure}[b]{0.02\linewidth}
	    \rotatebox[origin=t]{90}{\scriptsize Standard transformer}\vspace{2.8\linewidth}
	\end{subfigure}%
	\begin{subfigure}[t]{0.96\linewidth}
	    \centering
		\includegraphics[width=\cwidth,clip]{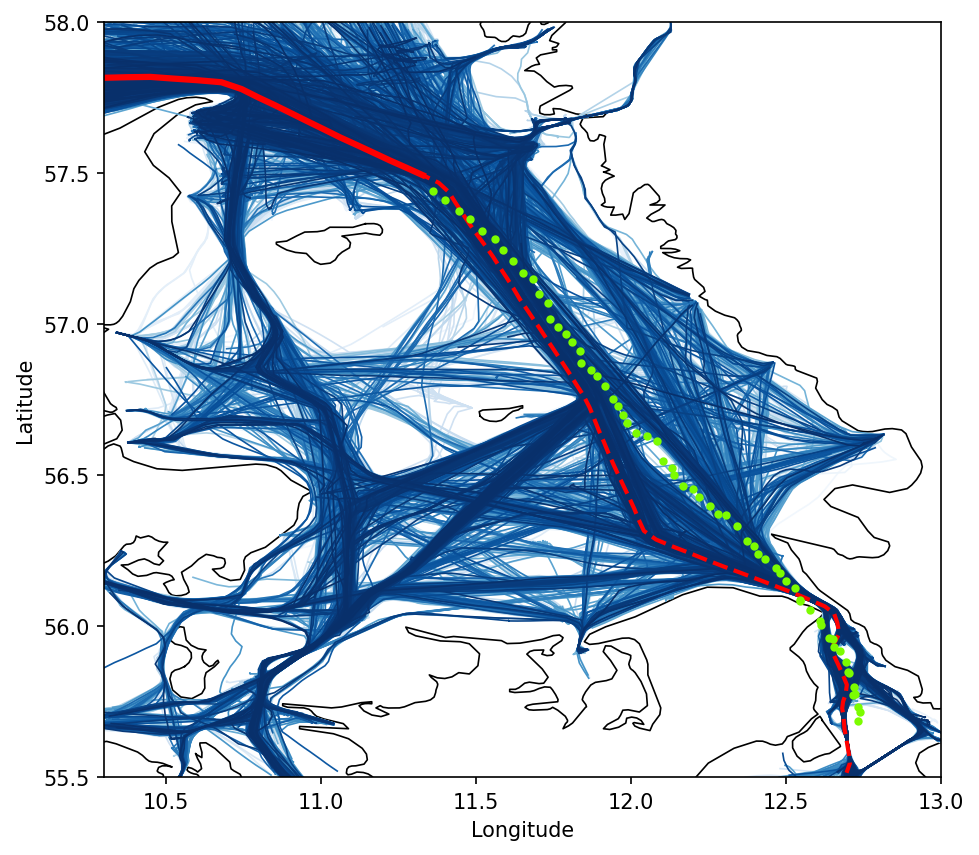}
		\includegraphics[width=\cwidth,clip]{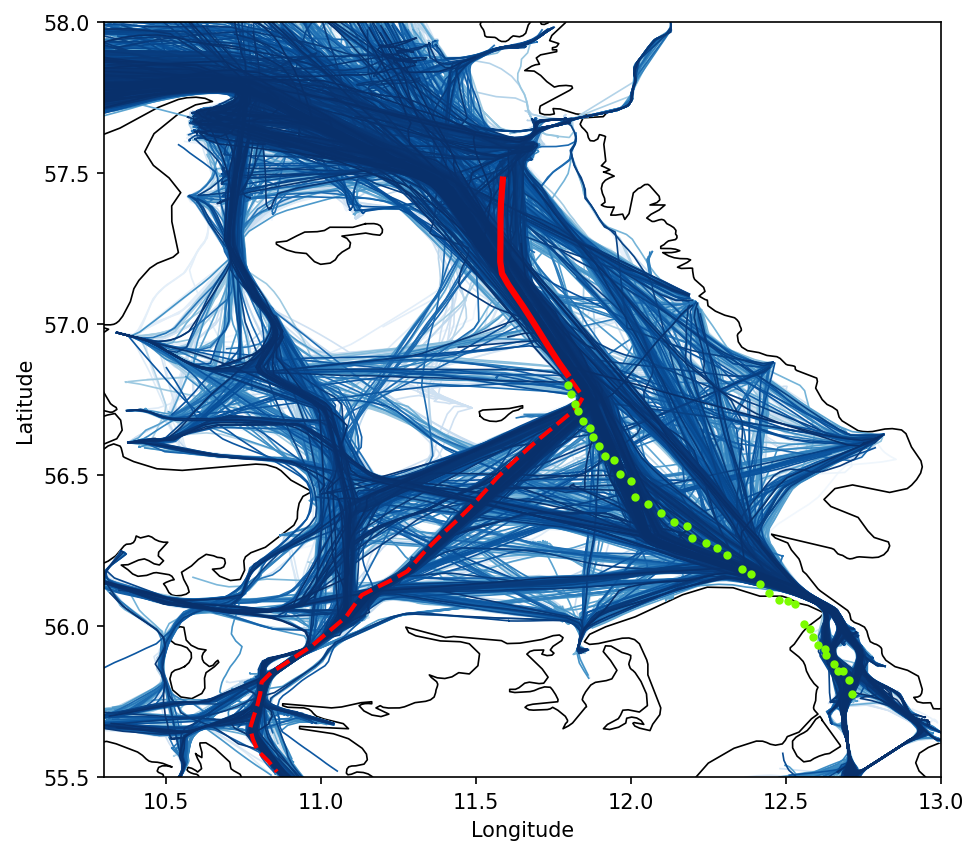}
		\includegraphics[width=\cwidth,clip]{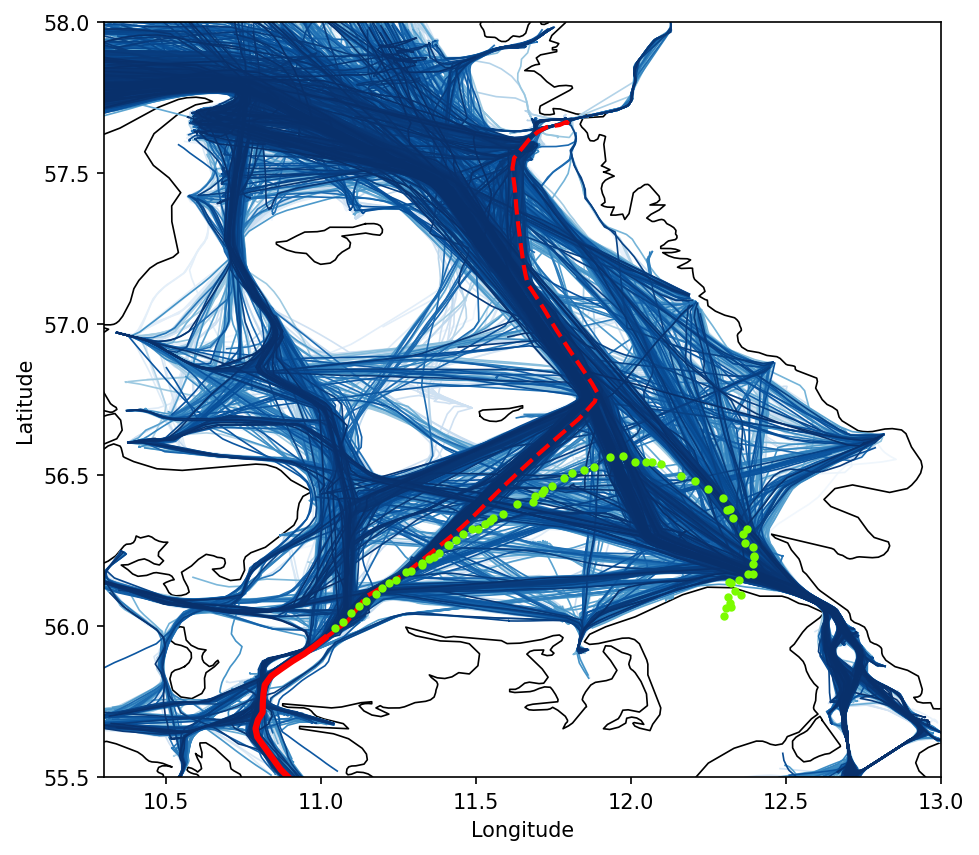}
		\includegraphics[width=\cwidth,clip]{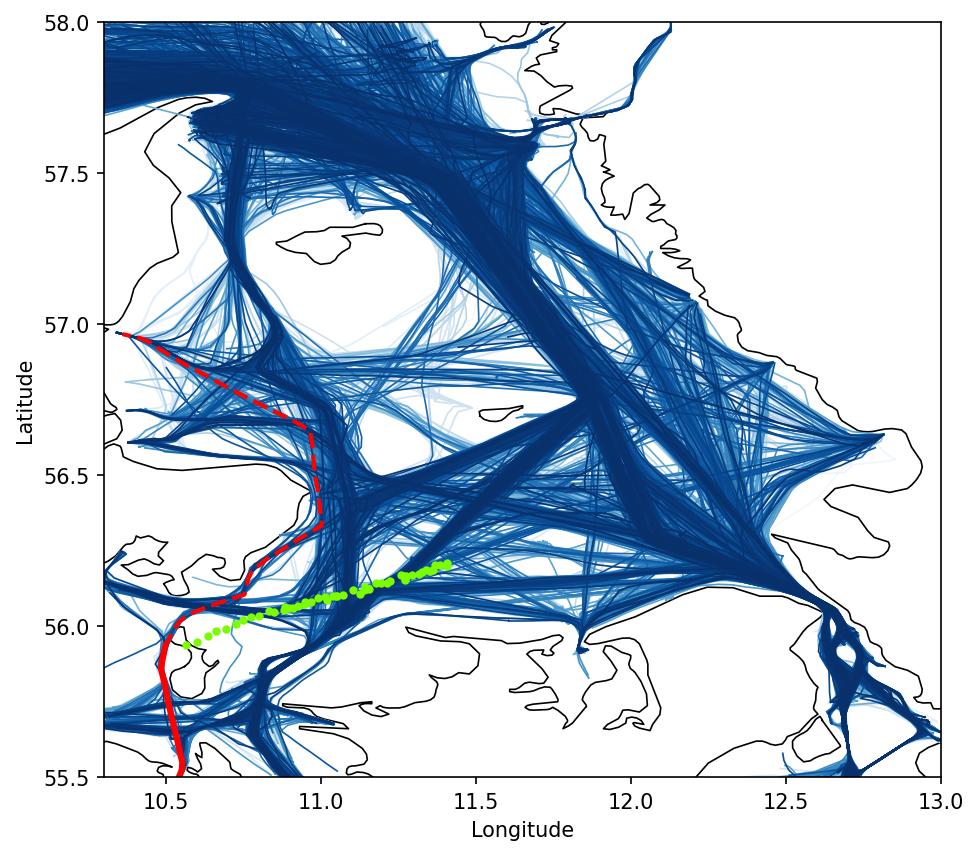}
	\end{subfigure}%
	
	\begin{subfigure}[b]{0.02\linewidth}
	    \rotatebox[origin=t]{90}{\scriptsize Without $\vect{h}_t$}\vspace{3.8\linewidth}
	\end{subfigure}%
	\begin{subfigure}[t]{0.96\linewidth}
	    \centering
		\includegraphics[width=\cwidth,clip]{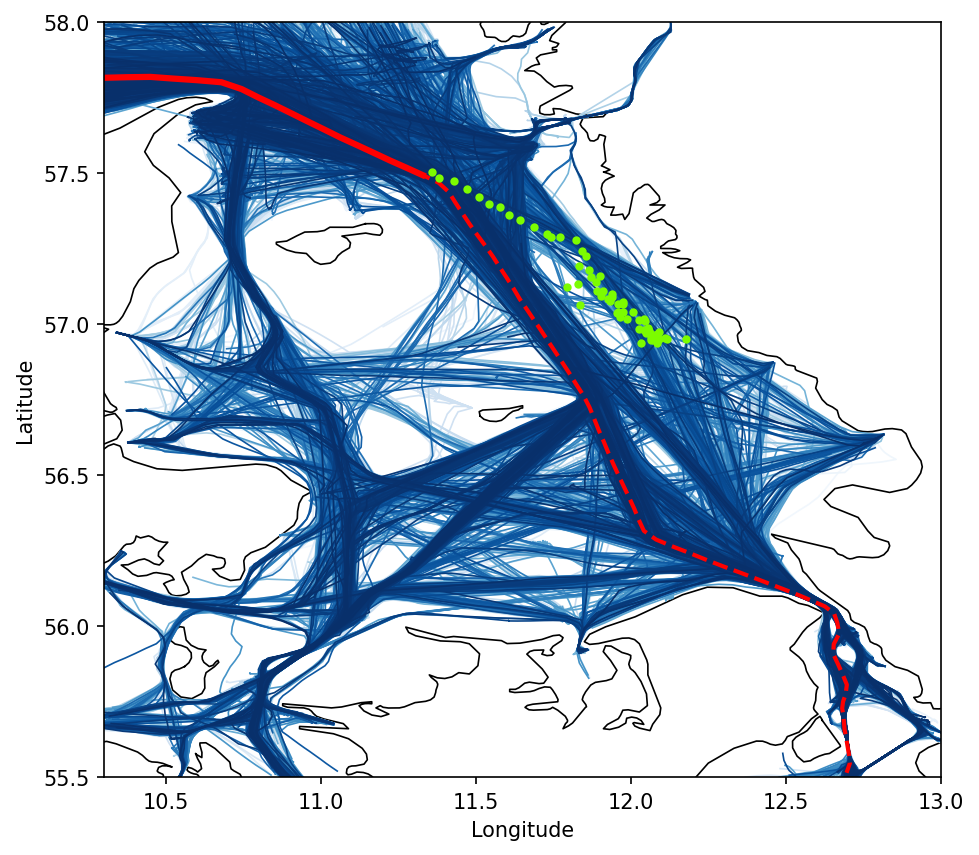}
		\includegraphics[width=\cwidth,clip]{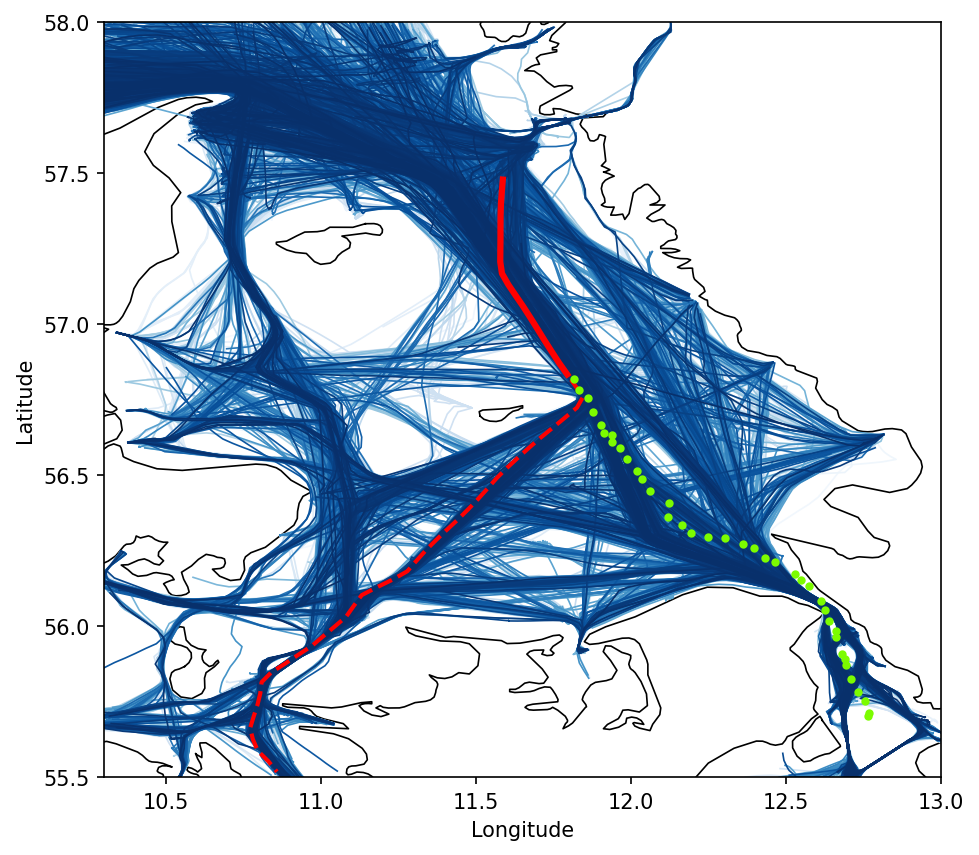}
		\includegraphics[width=\cwidth,clip]{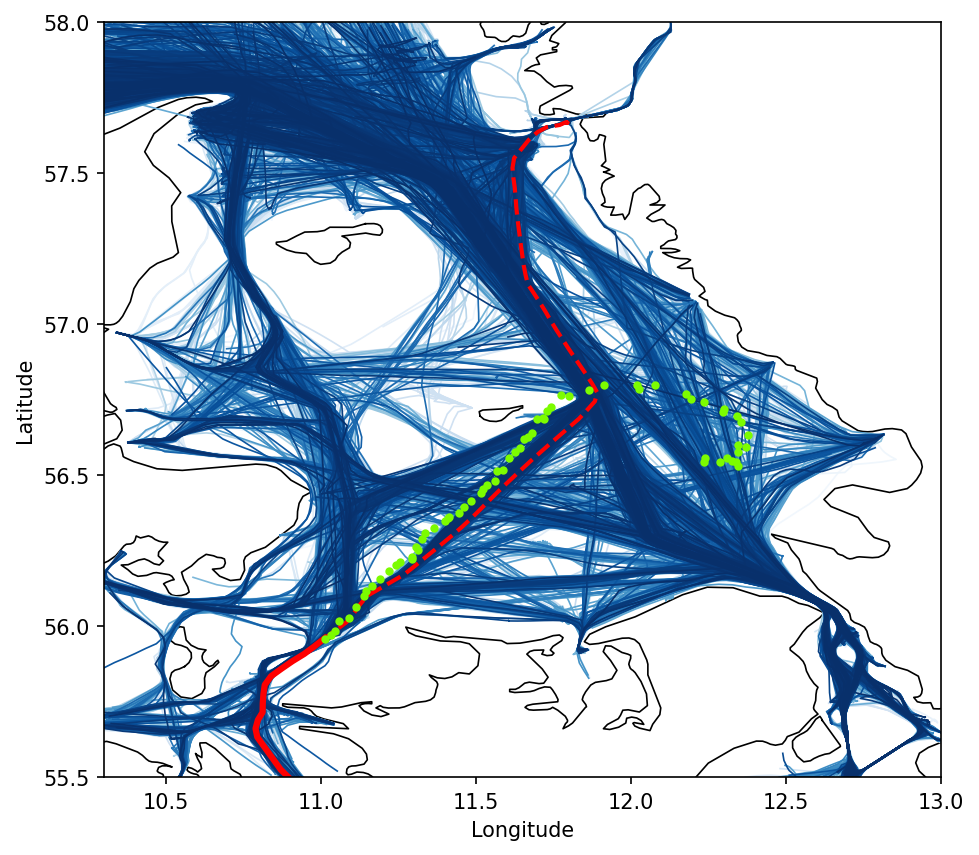}
		\includegraphics[width=\cwidth,clip]{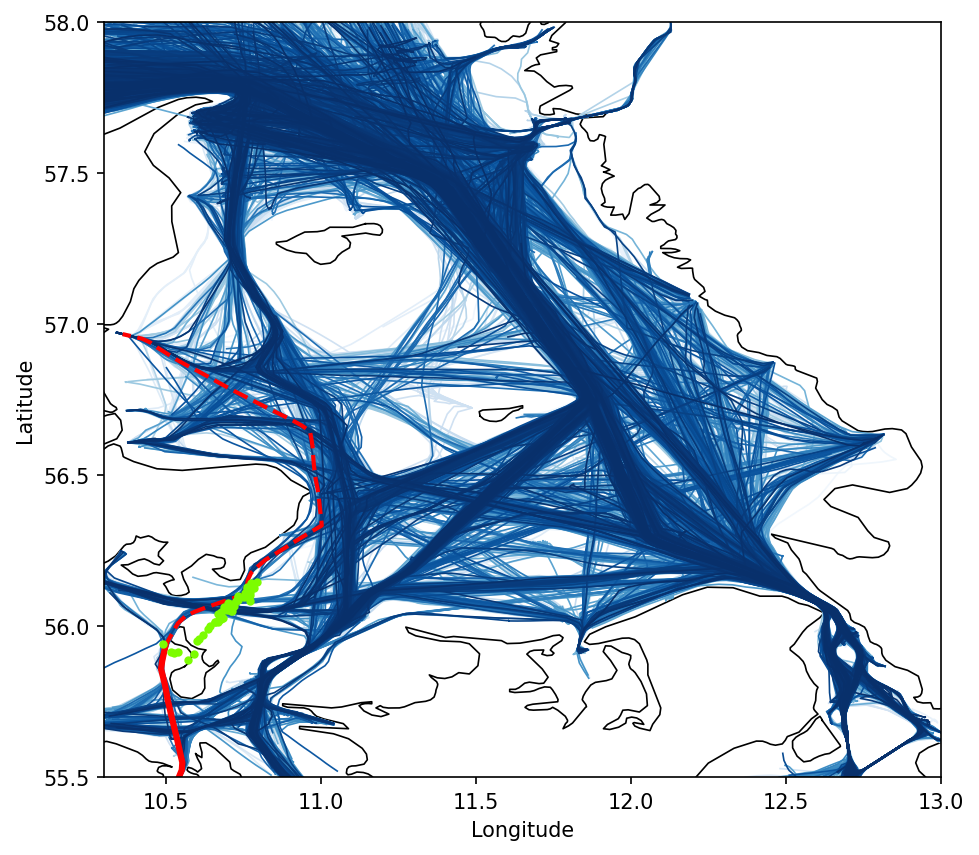}
	\end{subfigure}%
	
	\begin{subfigure}[b]{0.02\linewidth}
	    \rotatebox[origin=t]{90}{\scriptsize Without $\mathcal{L}_{CE}$}\vspace{3.8\linewidth}
	\end{subfigure}%
	\begin{subfigure}[t]{0.96\linewidth}
	    \centering
		\includegraphics[width=\cwidth,clip]{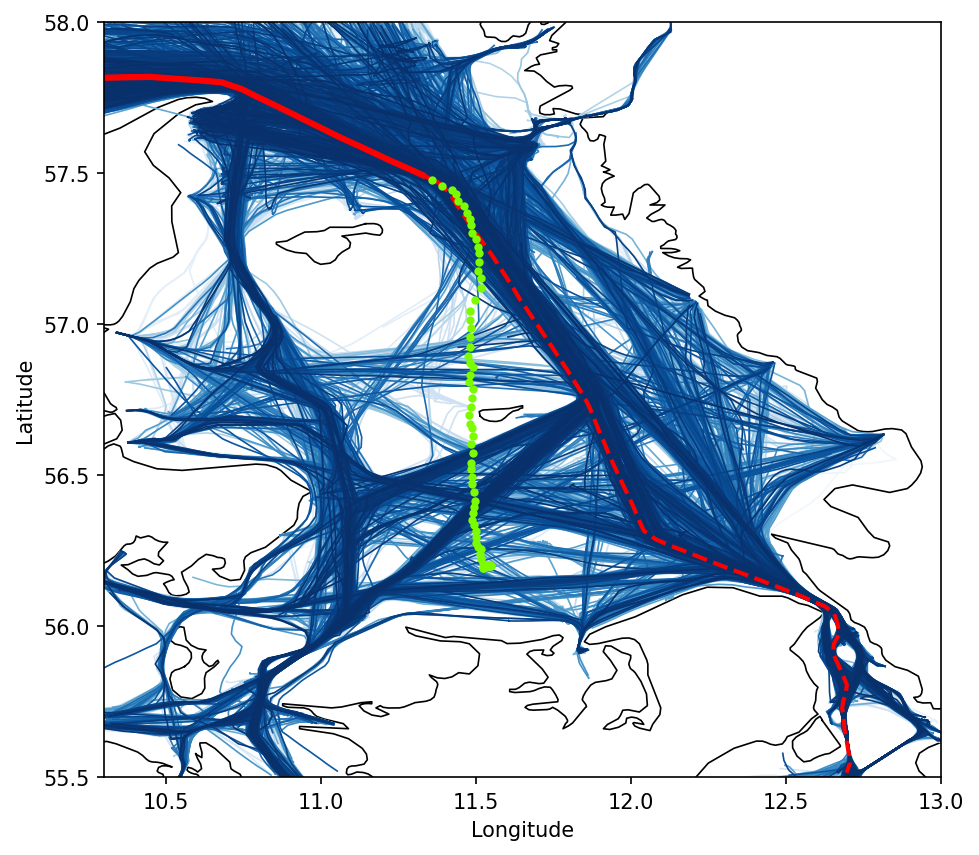}
		\includegraphics[width=\cwidth,clip]{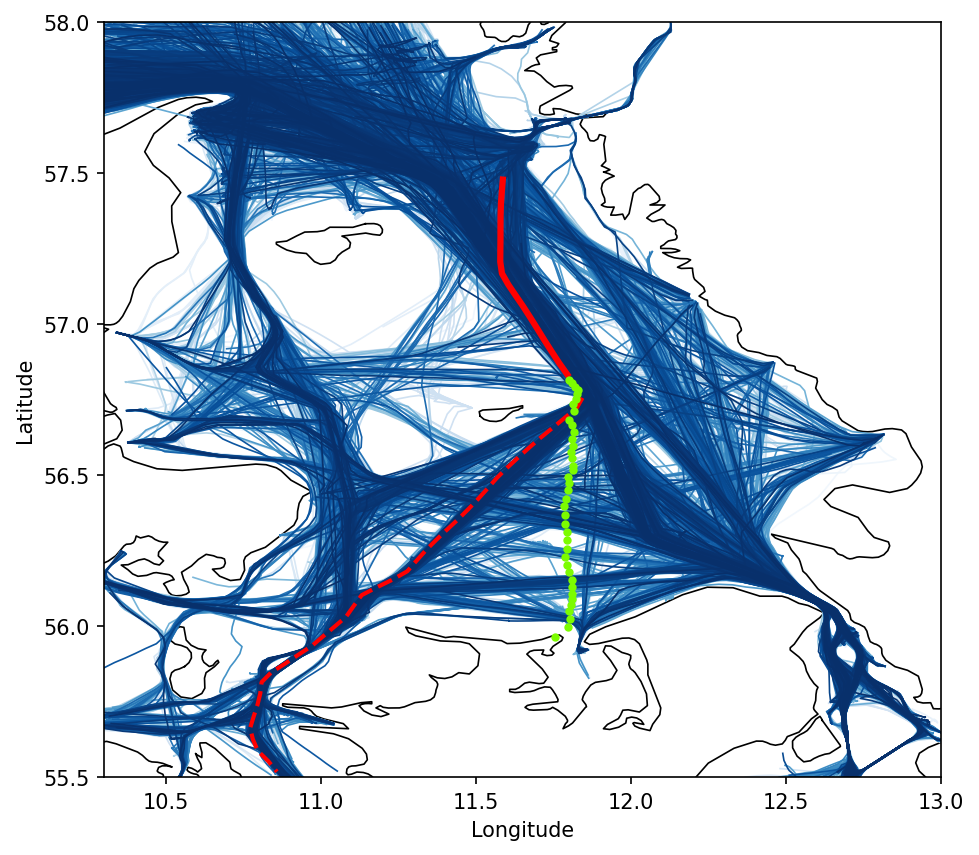}
		\includegraphics[width=\cwidth,clip]{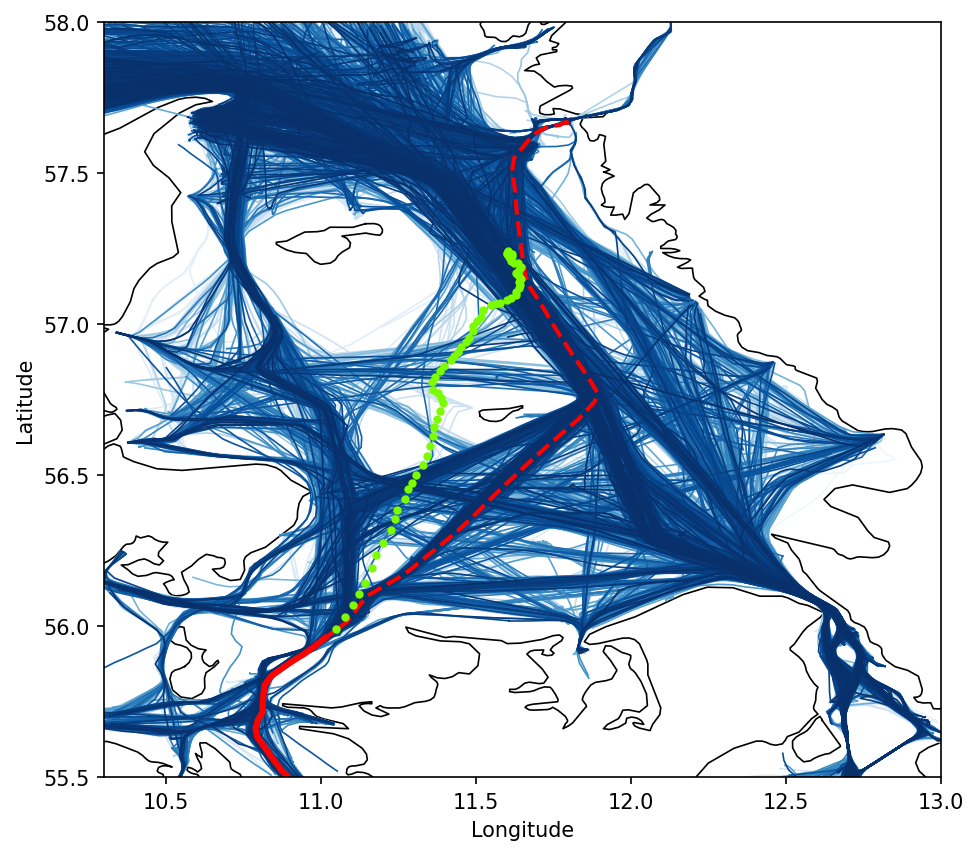}
		\includegraphics[width=\cwidth,clip]{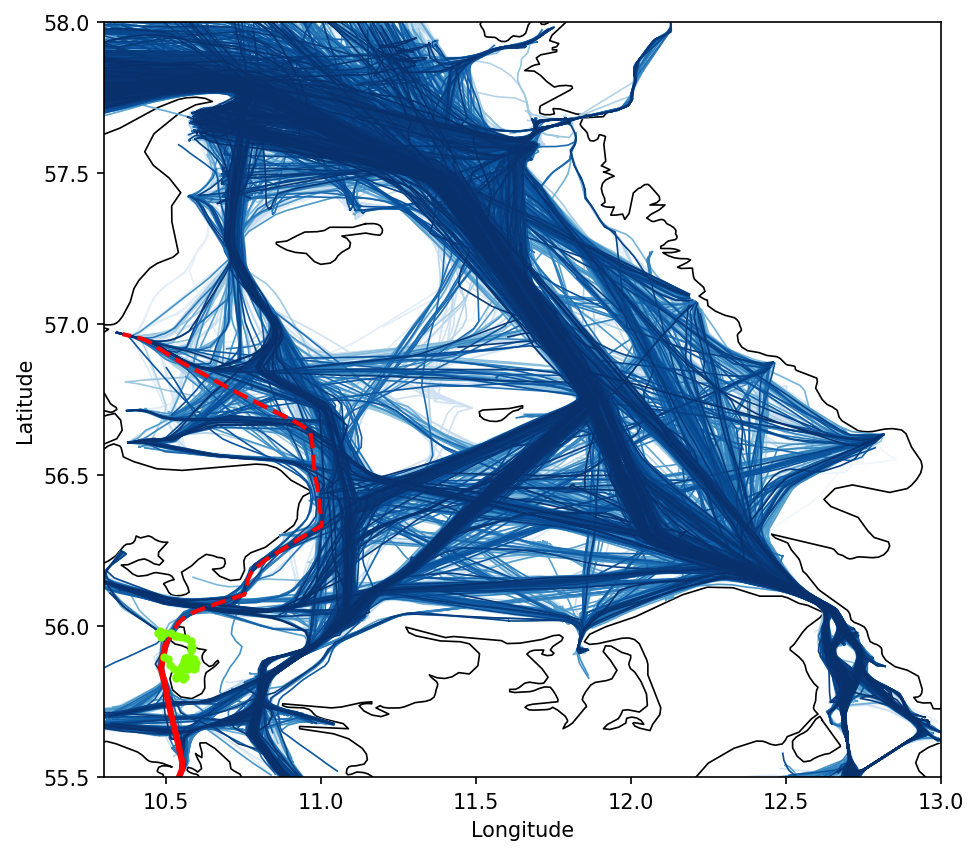}
	\end{subfigure}%

	\begin{subfigure}[b]{0.02\linewidth}
	    \rotatebox[origin=t]{90}{\scriptsize TrAISformer}\vspace{4\linewidth}
	\end{subfigure}%
	\begin{subfigure}[t]{0.96\linewidth}
	    \centering
		\includegraphics[width=\cwidth,clip]{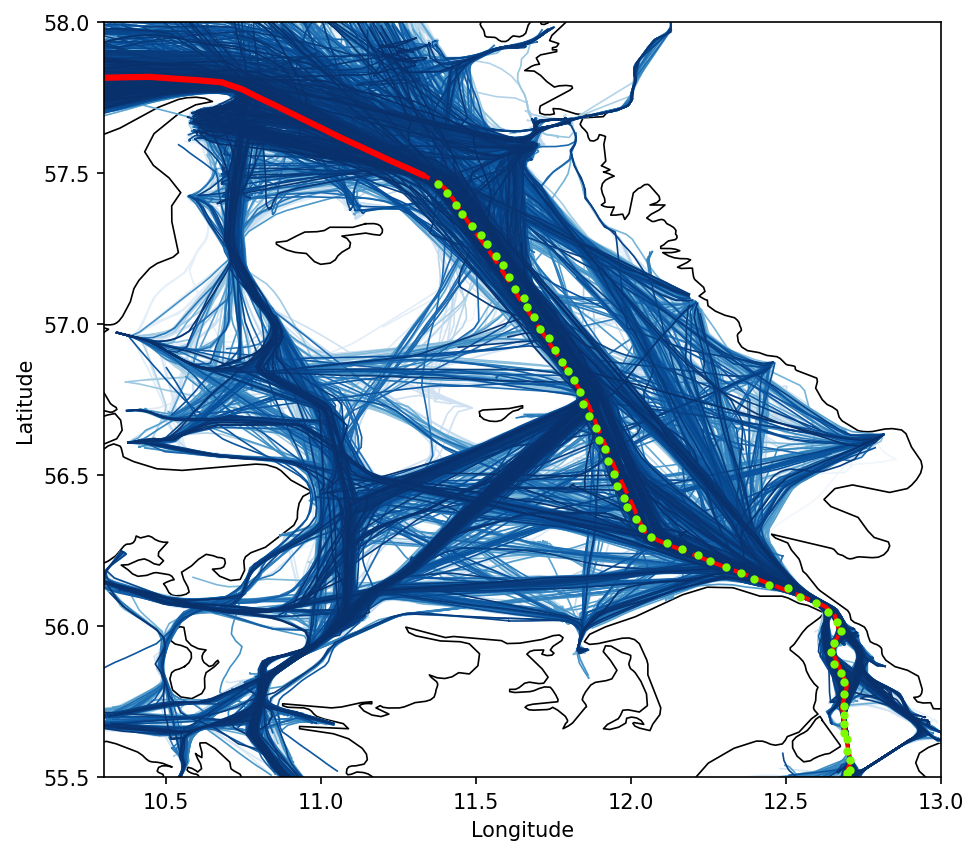}
		\includegraphics[width=\cwidth,clip]{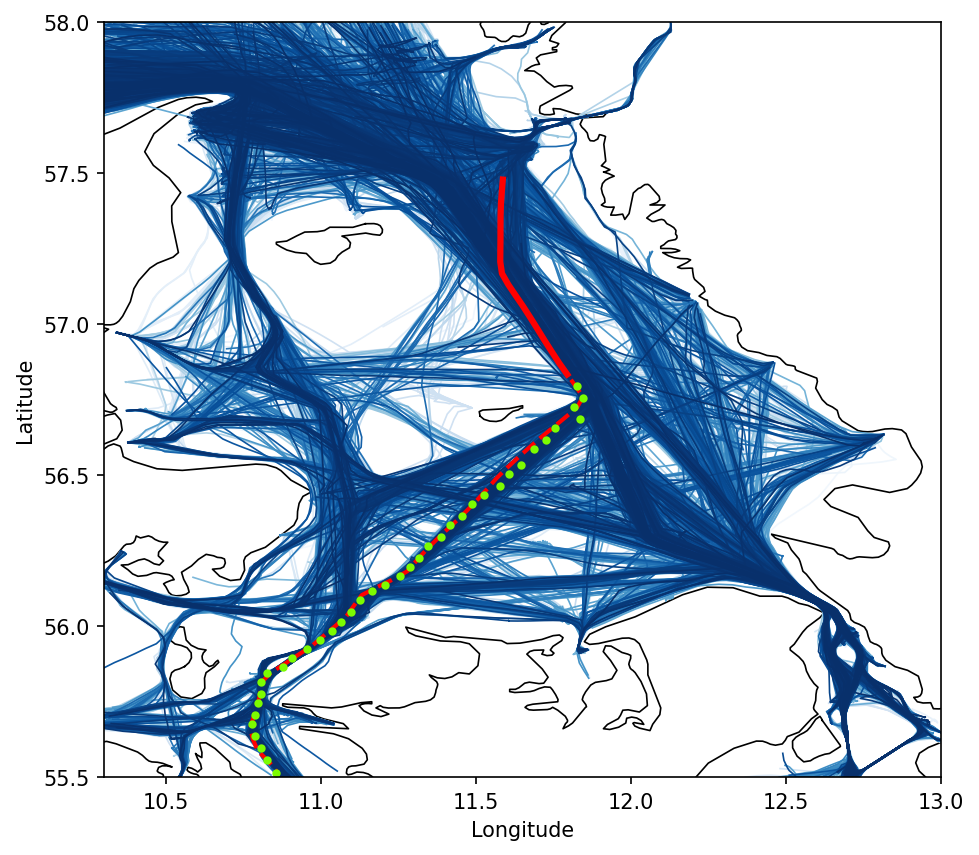}
		\includegraphics[width=\cwidth,clip]{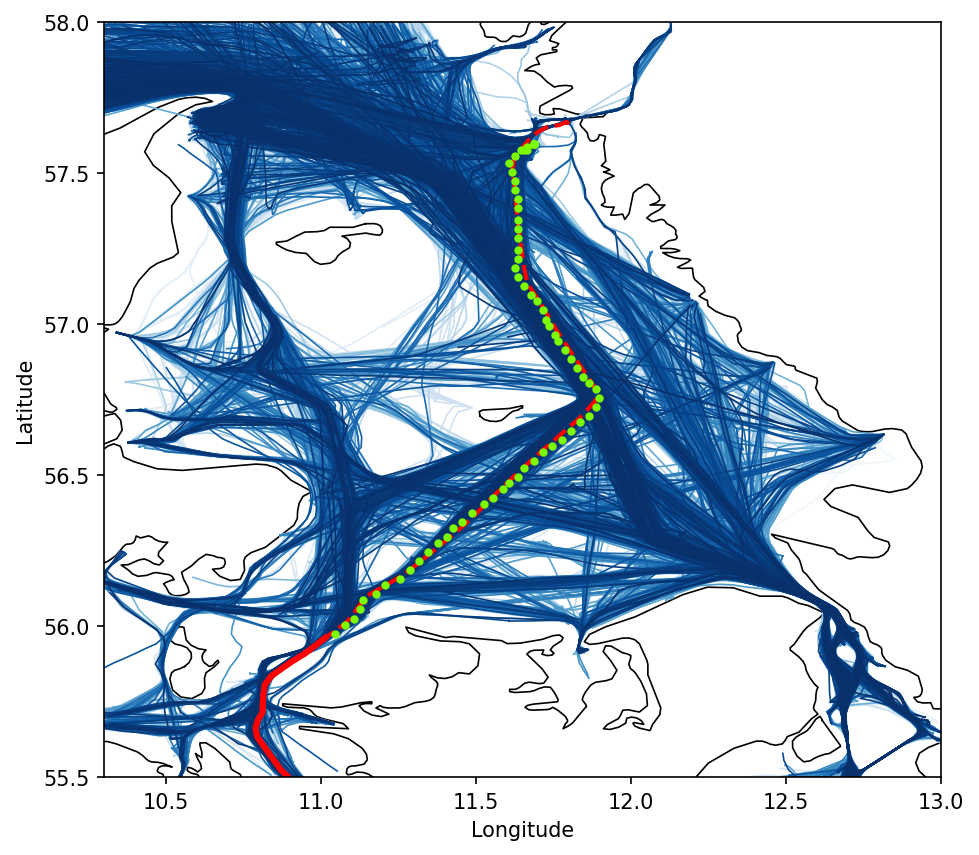}
		\includegraphics[width=\cwidth,clip]{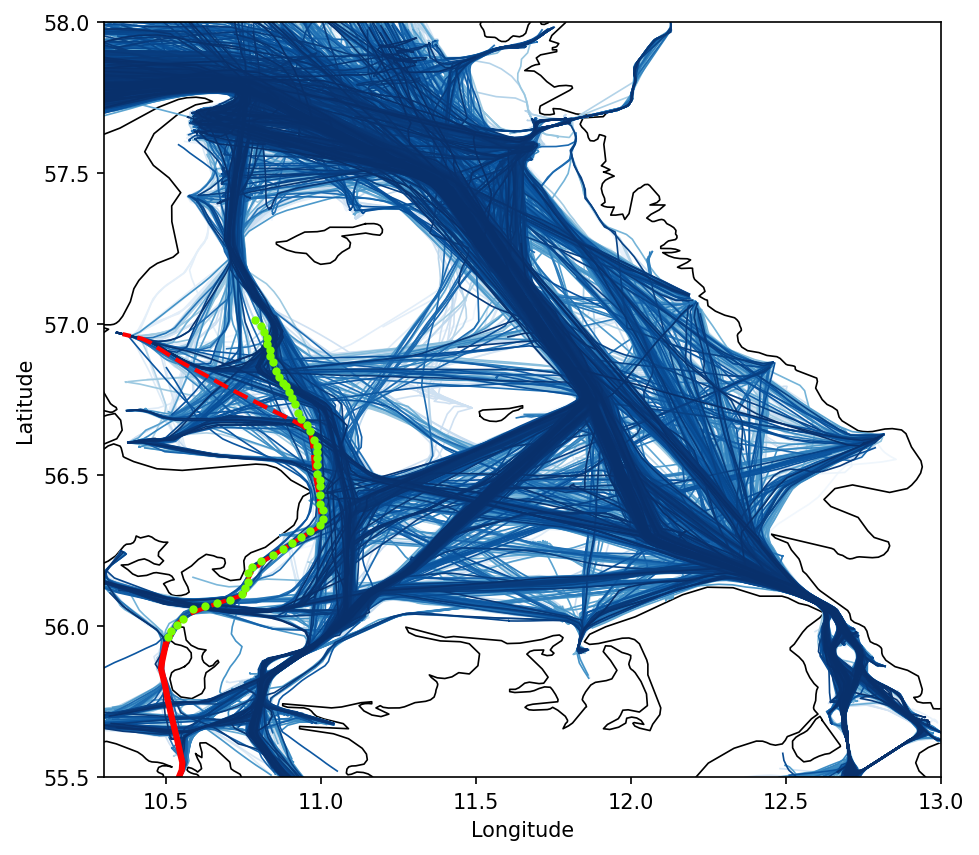}
	\end{subfigure}%
	
	\caption{\textbf{Examples of \rvv{AIS-based vessel trajectory} predictions}: each column depicts the predictions of a given real \rvv{vessel trajectory}. The rows correspond to the various models used for benchmarking. For each example, we display the AIS observations  $\vect{x}_{0:T}$ used as the input by all models {\color{red} \textbf{---}}, the real \rvv{vessel trajectory} {\color{red} \textbf{-}\textbf{-}} , and the predicted trajectory {\color{green} $\bullet$ }.} 
    \label{fig:predictions}
\end{figure*}

\begin{figure}
  \centering
  \includegraphics[width=0.9\linewidth]{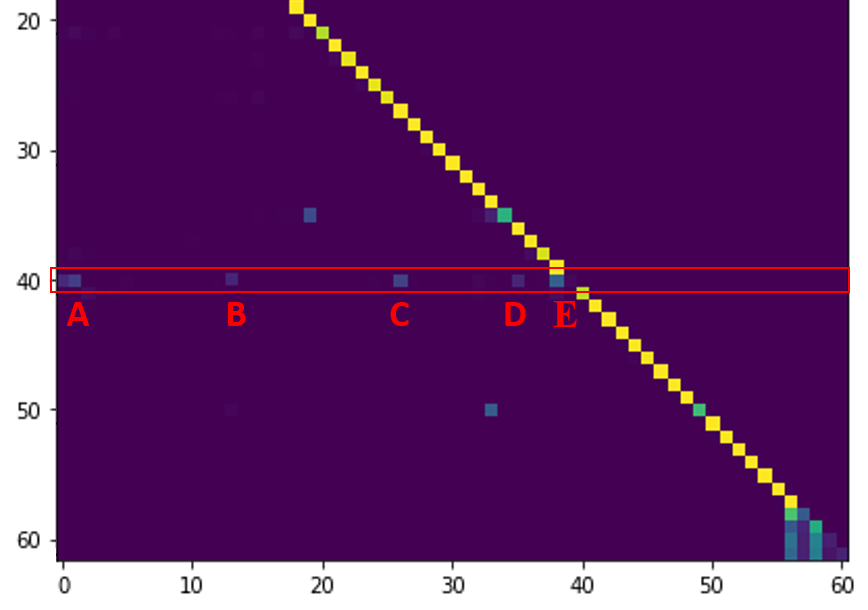}
  \centering
  \caption{\textbf{Relative importance of each time step in the prediction}: Visualization of the activation of one attention block of \textit{TrAISformer} for the trajectory shown in Fig. \ref{fig:maritmeTrafficGraph}. Horizontal axis: input time step; vertical axis: output time step. Each row shows which parts of the input that the model pays attention to in order to compute the output at the corresponding time step. } \label{fig:att}
\end{figure}

We further analyze in Fig. \ref{fig:att} the behavior of \textit{TrAISformer} through the activation of an attention block in the first layer of \textit{TrAISformer} for the trajectory shown in Fig. \ref{fig:maritmeTrafficGraph}. Each row shows the relative importance of each time step in the predicted output. Some remarks raised from this analysis: 
\begin{itemize}
    \item On straight lines, only the information from recent time steps is used to predict the next time step, which is similar to constant velocity models \cite{xiao_traffic_2020};
    \item At the waypoints, the model needs to retrieve information from much earlier time steps, especially at the previous waypoints to predict the next time step. For example, row 40 (the red rectangle) depicts the attention weights to compute the prediction at \textbf{E}. The model pays more attention to the inputs at \textbf{A}, \textbf{B}, \textbf{C}, \textbf{D}, and \textbf{E}. One intuitive explanation is that the model needs to know where the vessel comes from (points \textbf{A}, \textbf{B}, \textbf{C}), what the movement pattern of the vessel is in the current segment (point \textbf{D}), as well as the current position and velocity (point \textbf{E}) to guess the movement pattern to come. 
\end{itemize}         
As such, this example demonstrates the ability of \textit{TrAISformer} to extract relevant long-term dependencies to predict vessel trajectories. 

\subsection{Ablation study}

To evaluate the significance of the different components of \textit{TrAISformer} architecture, we conducted an ablation study:
\begin{itemize}
    \item Firstly, we removed $\vect{e}_t$ and $\vect{h}_t$ to demonstrate the significance of the high-dimensional encoding. This is equivalent to applying directly a GPT model \cite{radford_improving_2018} to 4-dimensional AIS data streams.
    \item Secondly, we kept $\vect{e}_t$ but removed $\vect{h}_t$ to assess the relevance of the sparsity  constraint. The embedding $\vect{x}_t \rightarrow \vect{e}_t$ in this model is a MultiLayer Perceptron (MLP).
    \item Finally, we tested a model with the same architecture as \textit{TrAISformer} but used a regression loss as the training loss to demonstrate the criticality of the classification loss.
\end{itemize}
The results in Tab. \ref{tab:ablation} show that all the ablated models lead to significantly worse performance compared with \textit{TrAISformer}. Interestingly, the performance degradation is in the same order of magnitude for the three ablated models, though the impact of the classification-based loss is slightly greater. Overall, these results highlight the importance of integrating all the components of our architecture for achieving the best prediction performance.

\begingroup
\begin{table*}[]
    \caption {{\bf Mean prediction performance (in nautical miles) of the models in the ablation study.}}
    \label{tab:ablation}
    \footnotesize
    \centering
    \begin{tabular}{l|c|c|c|ccc}
    \toprule
    Model & AIS data representation & Embedding $\vect{x}_t \rightarrow \vect{e}_k$ & Loss function & 1h & 2h & 3h \\

    \toprule
    Without $\vect{e}_t$ and $\vect{h}_t$ (standard transformer) & $\left[lat, lon, SOG, COG \right]^T$ & None & $\mathcal{L}_{MSE}$   & 4.75 & 8.36 & 11.40 \\

    \midrule 
    Without $\vect{h}_t$ & $\left[lat, lon, SOG, COG \right]^T$ $\rightarrow \vect{e}_t$ & MLP & $\mathcal{L}_{MSE}$ & 5.02 & 9.69 & 15.04 \\

    \midrule 
    Without the classification loss $\mathcal{L}_{CE}$ & ``four-hot`` vector $\rightarrow \vect{e}_t$ & Via $\vect{h}_t$ & $\mathcal{L}_{MSE}$ & 5.53 & 10.64 & 16.06 \\

    \midrule 
    \textit{TrAISformer} & ``four-hot`` vector $\rightarrow \vect{e}_t$ & Via $\vect{h}_t$ & $\mathcal{L}_{CE}$ & \textbf{0.48} & \textbf{0.94} & \textbf{1.64} \\

    \bottomrule
    \end{tabular}
    \normalsize
\end{table*}
\endgroup

\section{Conclusions and Future work}
\label{sec:conclusions}

In this paper, we presented a novel model---referred to as \textit{TrAISformer}, for vessel trajectory prediction using AIS data. The model uses an augmented, sparse, and high-dimensional representation of AIS data as well as a state-of-the-art transformer network architecture to learn complex patterns in vessel trajectories. 
Using a classification-based training loss, \textit{TrAISformer} can capture the multimodal nature of trajectory data. Experiments on real, public AIS data show \textit{TrAISformer} outperforms existing methods by a significant margin. With a 9-hour-ahead prediction error below 10 nmi on a real AIS dataset in a case-study region involving dense and complex maritime traffic patterns, these results open new research avenues for various applications such as search and rescue, port congestion avoidance, and maritime surveillance.

Through an ablation study, we have shown that all the above-mentioned components of \textit{TrAISformer} have critical roles in the reported performance. Though transformer architectures are likely not fully explainable \cite{adadi_peeking_2018, barredo_arrieta_explainable_2020}, we have also shown that the intermediate attention weights of the transformer architecture provide a natural way to explore how \textit{TrAISformer} exploits the past AIS data to compute its predictions of the future trajectory. This supports that the learned transformer representation could be of interest beyond the considered application to prediction tasks. 

Future work could further improve the architecture and develop the applications of \textit{TrAISformer} framework.  Among others, we may cite the learning of conditional \textit{TrAISformer} w.r.t. weather conditions as the latter clearly impact vessels' movement. While we currently omit the influence of vessel interactions, future work could study the possibility of integrating those interactions into the model. \rvvv{We may also stress that the proposed \textit{TrAISformer} architecture is significantly more complex (see Tab. \ref{tab:infomation_criteria}) with $\sim$300 times more parameters than the second most complex architecture among the benchmarked ones. While the greater complexity likely contributes to the significant gain, 
recent advances in model compression techniques, such as Neural Network Pruning \cite{molchanov_importance_2019} and Knowledge Distillation \cite{gou_knowledge_2021}, suggest that we could reduce the model's size typically by a factor of tens to hundreds without compromising its performance. This would promote the assessment and adoption of  \textit{TrAISformer} in operational systems.} The combination of  \textit{TrAISformer} with other learning-based modules for classification and anomaly detections \cite{nguyen_geotracknet-maritime_2021} is also of interest. Recent advances in the exploitation of AIS data for the inversion of sea surface parameters \cite{benaichouche_unsupervised_2021, benaichouche_unsupervised_2021-1} may also be an appealing line of research for our future work. 

\appendix
\section{Why pedestrian and vehicle trajectory prediction models do not apply to medium-range AIS-based vessel trajectory prediction}
\label{sec:short-term_long-term}

\rvv{In this appendix, we present a mathematical demonstration explaining why pedestrian and vehicle trajectory prediction models are not directly applicable to medium-range AIS-based vessel trajectory prediction.
}

\rvv{
Let us denote by $\vect{x}^{s^i}_t$ an observation of an agent $s^i$ at time $t$. For instance, in pedestrian and vehicle trajectory prediction, $s^i$ can represent either a pedestrian or a vehicle, and $\vect{x}^{s^i}_t$ corresponds to their respective positions on the map.  In AIS trajectory prediction, $s^i$ represents a vessel, and $\vect{x}^{s^i}_t$ represents its AIS message. The trajectory of agent $s^i$ from $t_1$ to $t_2$ ($t_2 > t_1$) is then represented by a sequence of observations $\vect{x}^{s^i}_{t_1:t_2} = \{\vect{x}^{s^i}_{t_1}, \vect{x}^{s^i}_{t_1+1},..., \vect{x}^{s^i}_{t_2}\}$. At time $t$, we denote the group of other agents in the vicinity of $s^i$ as $\vect{V}^i_t$, and their historical trajectories are denoted as $\vect{x}^{\vect{V}^i_t}_{0:t}$. 
}

\rvv{
In the context of this paper, using these notations, trajectory prediction refers to forecasting the trajectory of an agent $s^i$ for $L$ timesteps ahead, based on the historical observations of this agent and the others in the vicinity up to time $T$, by maximizing the likelihood:
}

\begin{equation}
    p(\vect{x}^{s^i}_{T+1:T+L}|\vect{x}^{s^i}_{0:T}, \vect{x}^{\vect{V}^i_t}_{0:T}).
    \label{eq:obj_basic_general}
\end{equation}
\rvv{
Here we use $p$ in the broad sense, which includes deterministic models. 
}
\begin{figure}
  \centering
  \includegraphics[width=0.9\linewidth]{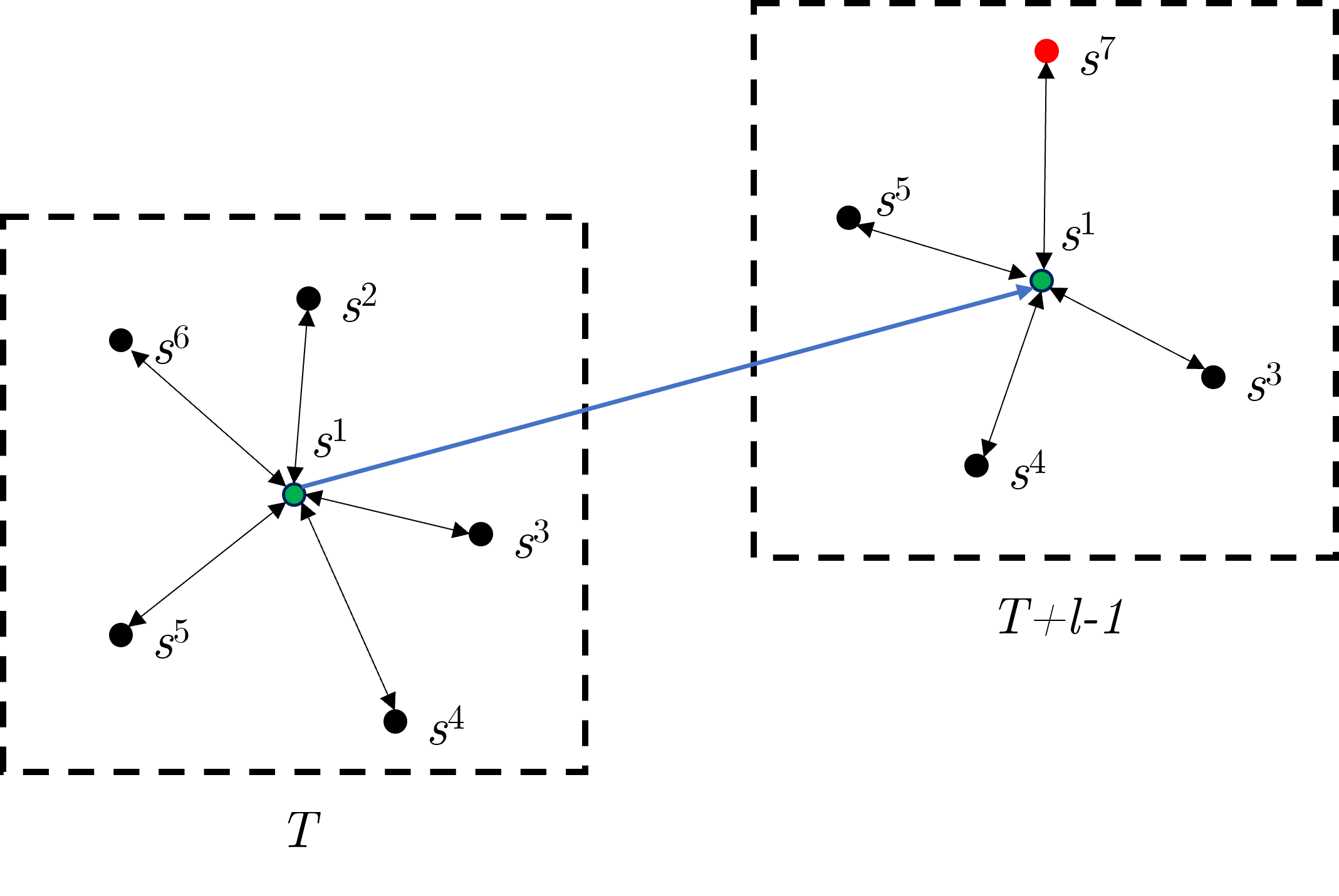}
  \centering
  \caption{\rvv{\textbf{Illustration of the intractability of the modeling of the interactions between agents in medium-range trajectory prediction}. Consider the scenario where we aim to predict the medium-range trajectory of agent $s^1$ (denoted by the green dot). At time $T$, there are six other agents ($s^2,...,s^6$) within its vicinity (enclosed by the dashed rectangle). The interactions between $s^1$ and these agents are depicted by the double-headed arrows. As we project into the medium-range future at time $T+l-1$, $s^2$ and $s^6$ will have moved away from $s^1$; and a new, unknown agent $s^7$ (denoted by the red dot) may appear near $s^1$. The prediction model would have no knowledge of this new vicinity, making the modeling of interactions between agents intractable.}} \label{fig:vicinity}
\end{figure}

\rvv{ The conditioning side of \eqref{eq:obj_basic_general} encompasses two crucial components: the $\vect{x}^{s^i}_{0:T}$ component embeds the intention of the agent, while the $\vect{x}^{\vect{V}^i_t}_{0:T}$ part embeds the interactions with the environment. State-of-the-art methods for pedestrian and vehicle trajectory prediction are centered on effectively modeling and integrating these two terms. Prominent examples include S-LSTM \cite{gupta_social_2018}, S-GAN \cite{gupta_social_2018}, S-ATTN \cite{vemula_social_2018}, SoPhie \cite{sadeghian_sophie_2019}, MATF \cite{zhao_multi-agent_2019}, Trajection \cite{ivanovic_trajectron_2019}, Trajectron++ \cite{salzmann_trajectron_2021}, etc. Those models leverage deep neural networks such as LSTM or GAN (Generative Adversarial Network) to capture correlations in historical data, and some pooling techniques to embed the interactions between agents. This idea has been adopted for short-term AIS-based vessel trajectory prediction \cite{liu_deep_2022, liu_stmgcn_2022}. Although those methods have shown promising results on the corresponding datasets, they are not suitable for medium-range vessel trajectory prediction. First, the prediction horizons considered in those works are from a few seconds to a few minutes, which are too short for maritime applications. Second, those works address different types of maneuvers, of which the movement of an agent depends highly on the interactions with other agents and the surrounding environment. For maritime traffic contexts, and at medium-range time horizons, the path that a vessel will make depends mainly on where it wants to go. It is barely affected by the interactions with other vessels in the vicinity at the current moment. Mathematically, this means at time $T$ we have the approximation:
}

\begin{equation}
    \left.{p(\vect{x}^{s^i}_{T+l}|\vect{x}^{s^i}_{0:T}, \vect{x}^{\vect{V}^i_t}_{0:T}) \approx p(\vect{x}^{s^i}_{T+l}|\vect{x}^{s^i}_{0:T}})\right\vert_{l \texttt{>>} 1}.
\end{equation}

\rvv{
One may argue that we could use the predicted value of $\vect{x}^{\vect{V}^i_{T+l-1}}_{0:T+l-1}$ and $\vect{x}^{s^i}_{0:T+l-1}$ to estimate $\vect{x}^{s^i}_{T+l}$. However, in order to predict $\vect{x}^{\vect{V}^i_{T+l-1}}_{0:T+l-1}$, we need to predict the trajectory of all the agents in the vicinity of $s^i$ at $T+l-1$. This is an expensive or even intractable approach. For example, an unknown agent may join the ROI, as illustrated in Fig. \ref{fig:vicinity}).
}

\rvv{
It's important to note that if we remove $\vect{x}^{\vect{V}^i_t}_{0:T}$ from \eqref{eq:obj_basic_general}, this objective function simplifies \eqref{eq:obj_basic}, which is the objective function used in medium-range AIS-based vessel trajectory prediction. Likewise, if we remove the module that encodes the interactions between agents in some of the pedestrian and vehicle trajectory prediction models mentioned above, we get models that have similar architectures to those designed for AIS trajectory prediction. For example, if we remove the interactions between agents part in Trajectron \cite{ivanovic_trajectron_2019}, it becomes an LSTM\_seq2seq model. 
}

\def\url#1{}
\bibliographystyle{IEEEtran}
\footnotesize
\vspace{-1.5em}
\bibliography{references}


\end{document}